\documentclass[conference]{IEEEtran}
\usepackage{cite}
\usepackage{amsmath,amssymb,amsfonts}
\usepackage{algorithmic}
\usepackage{graphicx}
\usepackage{textcomp}
\usepackage{placeins}
\usepackage{floatrow}
\usepackage{float}
\usepackage{subfig}
\usepackage{xcolor}

\def\BibTeX{{\rm B\kern-.05em{\sc i\kern-.025em b}\kern-.08em
    T\kern-.1667em\lower.7ex\hbox{E}\kern-.125emX}}
\newtheorem{defn}{Definition}
\begin{document}

\title{Simulate forest trees by integrating L-system and 3D CAD files}
% Leveraging 
%An effective way to comprehend of forest trees with mixed information on L-sys and 3D developed models 
%{\footnotesize \textsuperscript{*}Note: Sub-titles are not captured in Xplore and
%should not be used}
%\thanks{Identify applicable funding agency here. If none, delete this.}
%}
\author{\IEEEauthorblockN{M. Hassan Tanveer}
\IEEEauthorblockA{\textit{Dept. of Statistics} \\
\textit{Virginia Tech}\\
Blacksburg, USA \\
hassantanveer@vt.edu}
\and
\IEEEauthorblockN{Antony Thomas}
\IEEEauthorblockA{\textit{DIBRIS} \\
\textit{University of Genova}\\
Genova, Italy\\
antony.thomas@dibris.unige.it}
\and
\IEEEauthorblockN{Xiaowei Wu}
\IEEEauthorblockA{\textit{Dept. of Statistics} \\
\textit{Virginia Tech}\\
Blacksburg, USA \\
xwwu@vt.edu}
\and
\IEEEauthorblockN{Hongxiao Zhu}
\IEEEauthorblockA{\textit{Dept. of Statistics} \\
\textit{Virginia Tech}\\
Blacksburg, USA \\
hongxiao@vt.edu}
}

\maketitle

\begin{abstract}

%\textcolor{red}{This research strengthens Lindenmayer system's capabilities by using information from meshes and vertices to formally describe the foliage using 3D CAD-developed tools}. 
In this article, we propose a new approach for simulating trees, including their branches, sub-branches, and leaves. This approach combines the theory of biological development, mathematical models, and computer graphics, producing simulated trees and forest with full geometry. Specifically, we adopt the Lindenmayer process to simulate the branching pattern of trees and modify the available measurements and dimensions of 3D CAD developed object files to create natural looking sub-branches and leaves. Randomization has been added to the 
placement of all branches, sub branches and leaves. 
To simulate a forest, we adopt Inhomogeneous Poisson process to generate random locations of trees.  Our approach can be used to create complex structured 3D virtual environment for the purpose of testing new sensors and training robotic algorithms. We look forward to applying this approach to test biosonar sensors that mimick bats' fly in the simulated environment.

%In this article, an experimental evaluation to imitate trees, their branches, sub branches, and leaves for simulation environment has been done. The overall goal is to make a forest that looks natural in visualization environment by modeling trees. It is a method for creating virtual trees that combines biological knowledge, mathematical formalism and computer graphics. Recently, the most common technique used for modeling plants is the Lindenmayer process or L-system. In this research we have upgraded the L-system branching pattern and modified the modeling of trees using 3D CAD developed object file that are created by using the available measurements and dimensions. Further the randomization has been added in all branches, sub branches and leaves placements. The trees has been plotted in 3D environment and Inhomogeneous Poisson process (IPP) taken into account to generate random points for tree placement in simulation environment. 
\end{abstract}

\begin{IEEEkeywords}
L-systems, CAD, Inhomogeneous Poisson Processes (IPPs), simulated trees 
\end{IEEEkeywords}

\section{Introduction}

%Humanity had always been influenced by natural observations in design and development. 
Thousands of years of biological evolution ultimately resulted in complex structured natural environments which play an important role for the survival of a wide variety of species and ecosystems \cite{renouf2005understanding}. The modeling of natural environments with forest trees has many theoretical and practical applications \cite{monserud1976simulation,cote2018fine}, ranging from conceptual nature studies, to forest-based visual impact analysis and dynamic landscape synthesis for computer-based animation \cite{tversky2002animation}. The inherent difficulty of tree modeling can be handled by using different approaches. For example, pioneering research methods for the efficient modeling of plant architecture have become available since 1971 \cite{HONDA1971331,perttunen2005incorporating}.
%Most approaches in the literature deal with trees in forest. We offer a short description of the some of this works. 
More recently, Reffye et. al \cite{de1995model} modeled plant growth and architectures by using  stochastic processes combined with botanical analysis.
For the three-dimensional (3D) modeling of individual trees, a number of methods have been proposed, including rule-based modelling \cite{lintermann1996interactive}, interactive parametric modelling \cite{hamon2012rtil}, image-based 3D modelling \cite{santos2012image}, and 3D reconstruction based on plant structure data \cite{paproki2012novel}. %Hundreds of detailed models of trees can be found on the web, many of which are in the form of standardized geometries that were mainly implemented to game designs, films and landscapes and therefore can not be easily modified. 
%A few forest visualization systems have been developed, 
Software systems, such as the stand visualization system (SVS) \cite{mcgaughey1997visualizing}, the forest vegetation simulator (FVS) \cite{crookston2010addressing}, and Capsis \cite{dufour2012capsis}, have been developed for the modeling and visualization of forest trees.
%\href{http://forsys.cfr.washington.edu/svs.html}    forest vegetation simulation (FVS) \cite{crookston2010addressing}, Capsis \cite{dufour2012capsis}.

Despite these advances, existing tree models are still not sufficient for simulating the full geometry of forest trees. Tree models in available software often rely on plugins (for e.g: blender tree plugins). %While they can be implemented to games and films, they can not be easily modified. \cmt{(Hassan: is this what you mean?)}} 
%While many detailed tree models have been made availale on the web, many of which are in the form of standardized geometries that were mainly implemented to game designs, films and landscapes and therefore can not be easily modified.  
Moreover, simulating the full geometry of 
natural trees often involves far more parameters than those used in simplified tree models. Empirical tree models that are based on real tree templates are sometimes useful to construct natural looking trees. 
Additionally, due to the large variety of tree species, simulating tree structures of a large family of tree types is a challenging task. %\cmt{(I somehow rewrote your paragraph below; not sure if this is what you meant; feel free to change.)} 
%The tree model is poorly represented in most simulations of forests. The tree structure systems are highly complex and constructing a detailed model of real tree for scientists in all the related fields, i.e., biology and agronomy, is difficult. The geometric parameters of tree models are too numerous and do not correspond to the tree architectural parameters predicted by empirical forest growth models. Therefore, integrating these models still remains a challenge.

In this paper, we propose an efficient approach for simulating natural looking trees by combining probabilistic models with empirical tree models. In particular, we integrate the commonly adopted Lindenmayer systems (L-systems) with 3D CAD object files, producing random tree architectures that look natural. The L-system \cite{ijiri2006sketch} 
is a graphical model commonly used to define the branching patterns in trees and other organic forms \cite{prusinkiewicz1988developmental}. It defines the branching pattern through recursively applying
certain production rules on a string of symbols. Each
symbol in the string defines a structural component (e.g., branch, terminal). Each recursive iteration creates an additional level of growth of the string. The final string represents the branching structure of the grown tree.
 %By increasing the recursion levels in the L-systems tree, it forms realistic patterns as occurring in the nature. As a result it can be used to define the branching patterns in trees and other organic forms \cite{prusinkiewicz1988developmental}. %\cmt{(What is the logic? Do you mean that the work you reviewed in previous paragraph are mostly based on L-system? And how is this paragraph connects in logic to the next paragraph?)}
%Yet, simulation methods should be integrated into dynamic growth models to automate the study and interpretation of tree growth processes in a simulated forest environment. 
While L-systems are commonly used to produce branching structures, they are not sufficient for
generating natural looking trees in 3D because of over-simplified assumptions on the geometry of branches, sub-branches and leaves. We therefore further improve L-systems by including geometric structures available in 3D CAD developed object files. %This creates natural looking sub-branches and leaves through using
%3D CAD-developed tools. This results in upgrading the L-system branching pattern and modification in the modeling of trees using 3D CAD developed object file that are created by using the available measurements and dimensions. 
The randomization has been added in all branches, sub branches and leaves placements. This creates random trees with natural looks. To simulate a forest, we adopt Inhomogeneous Poisson process (IPP) to generate random locations of trees. The simulated forest is plotted in 3D for visualization. Our approach can be used to create complex structured 3D virtual environment for the purpose of testing new sensors and training robotic algorithms. We look forward to applying this approach to test biosonar sensors that mimick bats' fly in the simulated environment. 
%The trees has been plotted in 3D environment and Inhomogeneous Poisson process (IPP) has been considered in order to generate random points for tree placement in simulation environment. 

The rest of this paper is organized as follows. In Section II, we describe in detail the methods for integrating L-system with 3D developed CAD files for the simulation of random trees, and discuss the random placement of trees in forest by sampling from the IPP. We elaborate the simulation results in Section III. Finally, in Section IV, a general conclusion and direction towards future work is given.

%Yet, most drones fail to navigate effectively in cluttered environment, albeit equipped with the best of sensors. There exists a number of reasons that renders these drones unsuitable for heavily cluttered environment--- weight of the sensors and the drones which in turn affect the maneuverability, poor vision in low-lit conditions, maximum speed.  

\begin{figure*}[h!]
    \centering
    \subfloat{\includegraphics[scale=0.25]{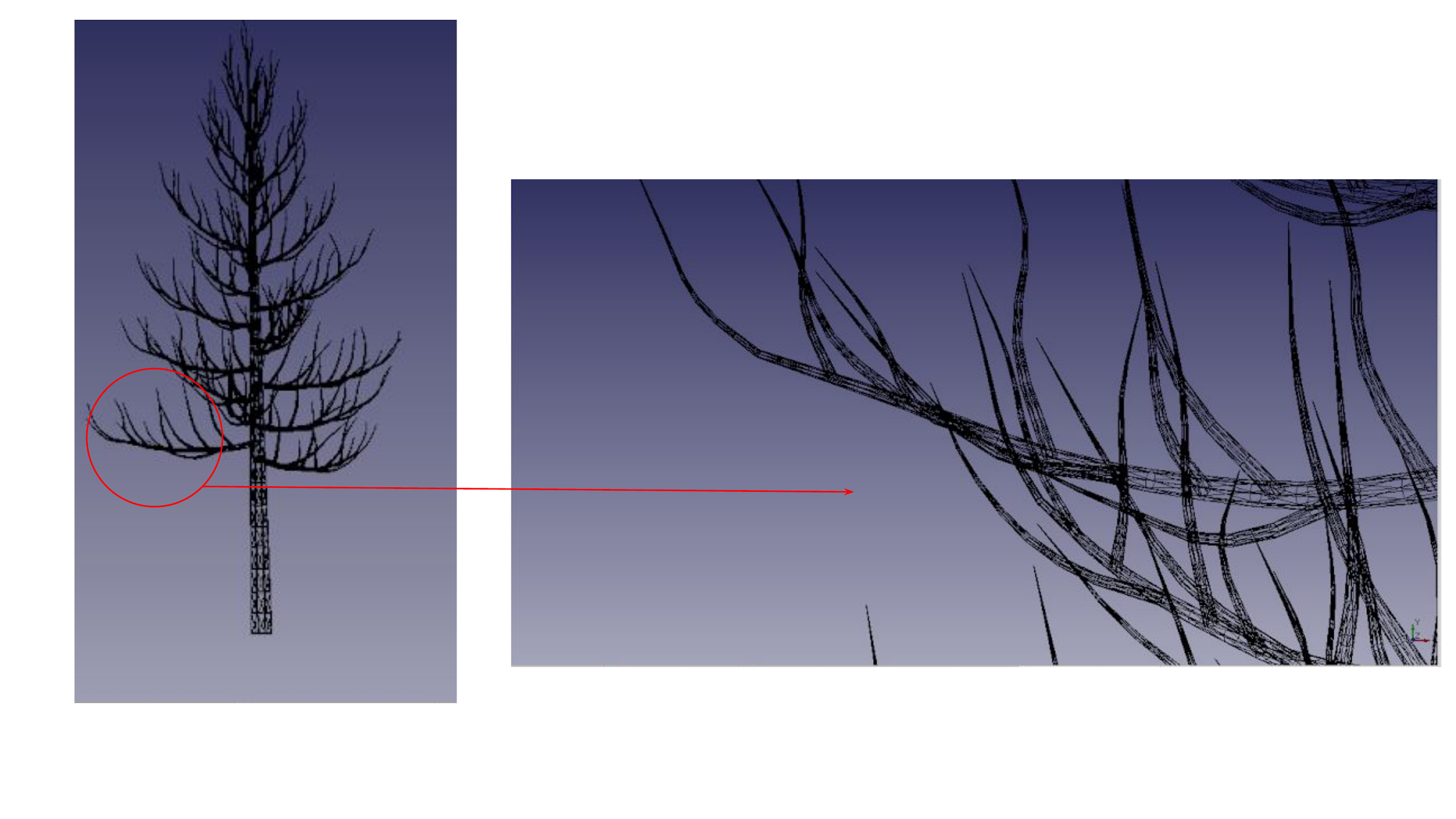}}
\subfloat{\includegraphics[scale=0.25]{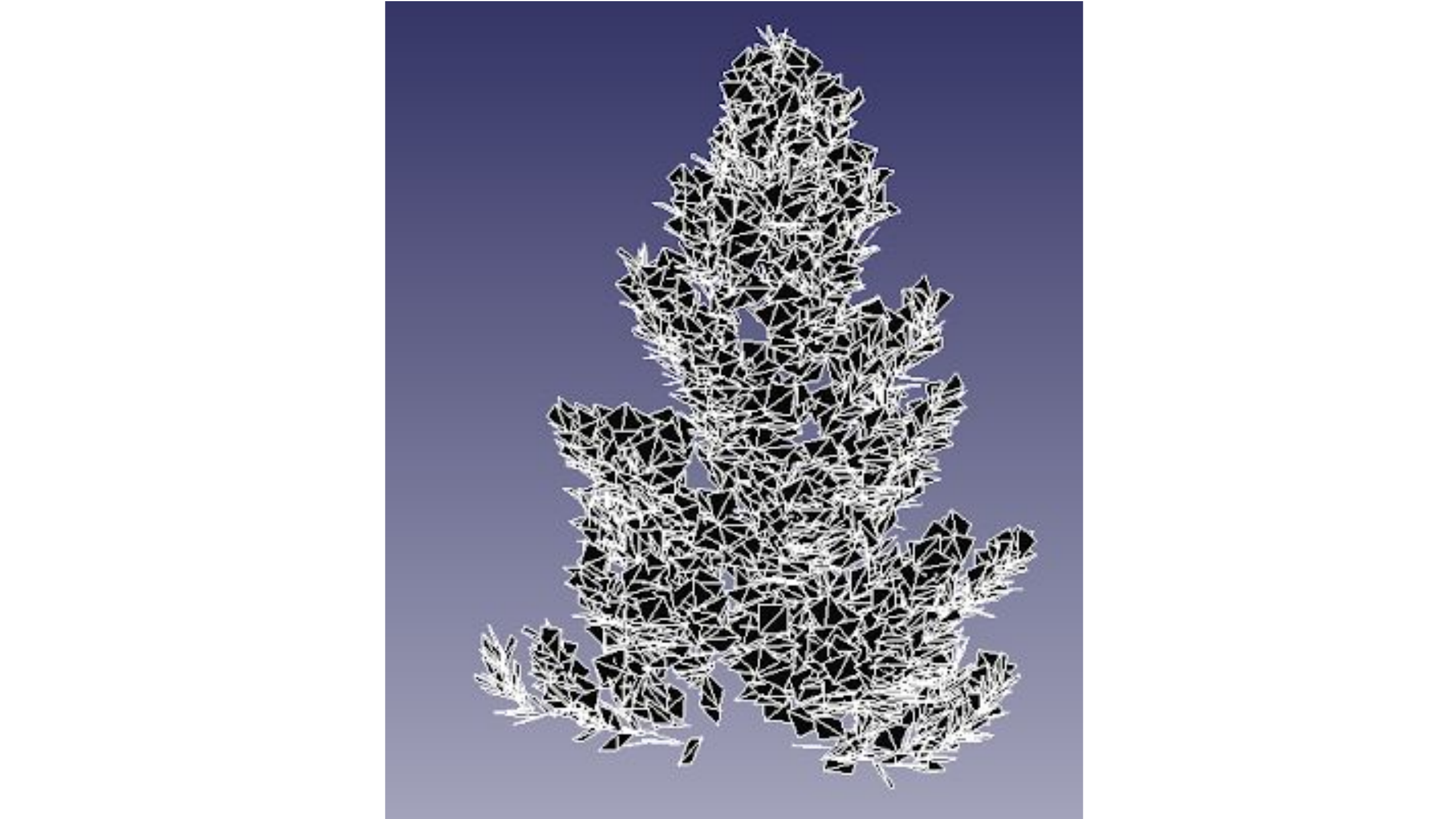}}
        \caption{Tree model visualized in a CAD viewer. (left) A 3D model of a pine tree with trunk and branches. Triangular meshes can be seen in the zoomed in view. (right) Tree is visualized with the leafs, which are modeled as triangular meshes.}
        \label{figcad1}
\end{figure*}

\begin{figure*}[h!]
    \centering
    \subfloat{\includegraphics[scale=0.25]{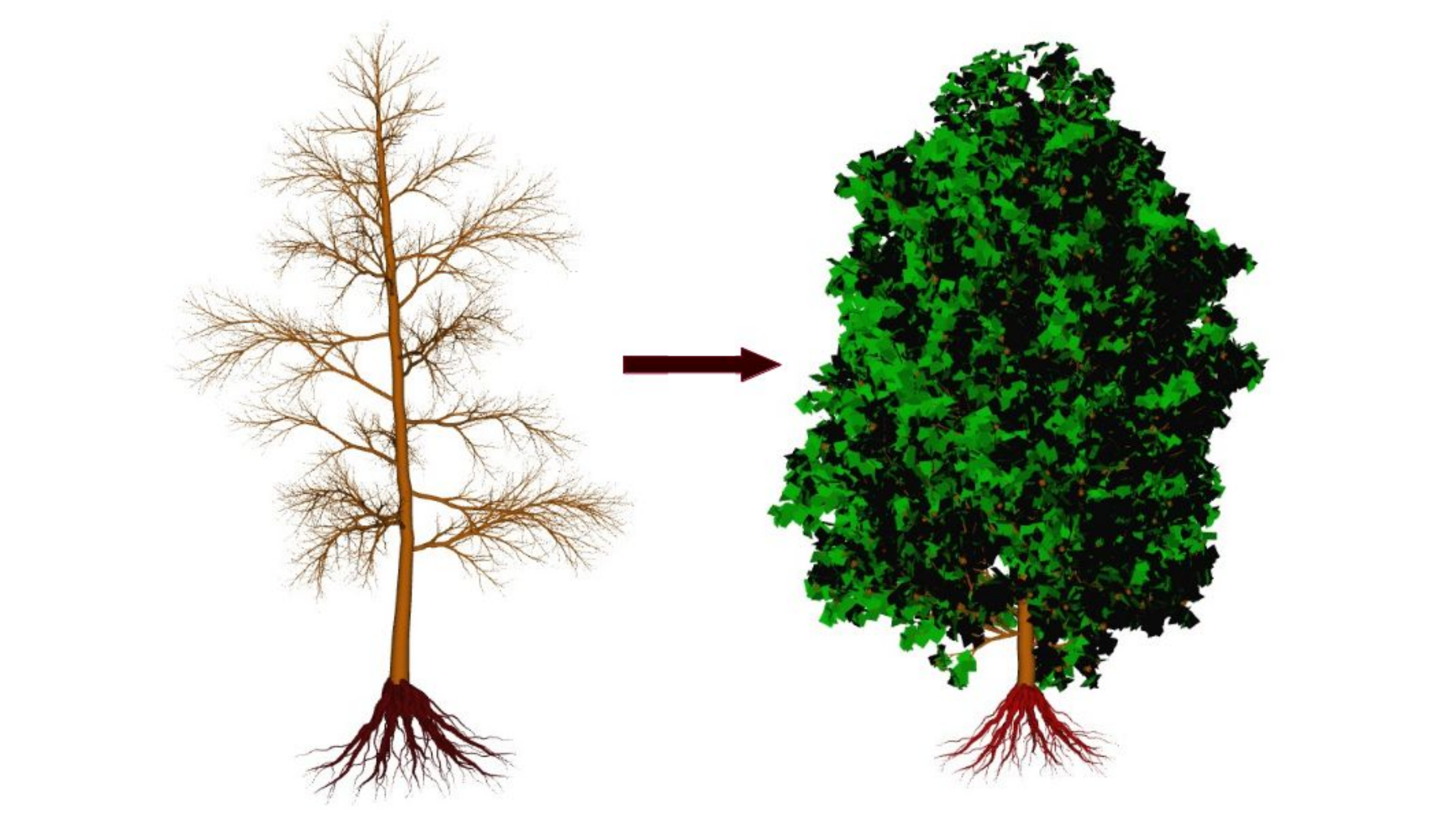}}
\subfloat{\includegraphics[scale=0.25]{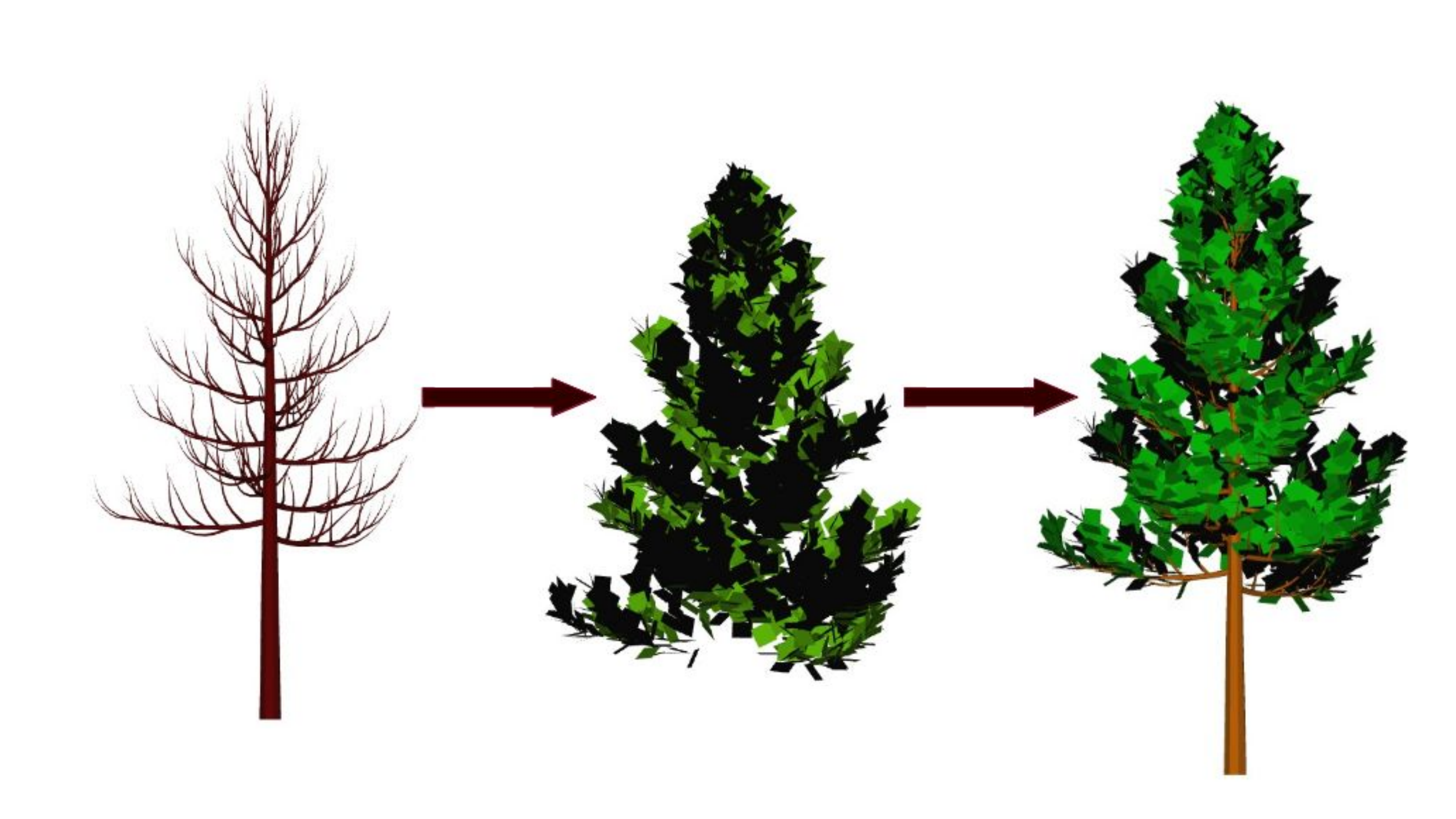}}
        \caption{STL files displayed as 2D images. 2D images are generated by 3D rendering or visualization of the corresponding STL files. }
        \label{figcad2}
\end{figure*}

\section{Materials and Methods}

\subsection{The L-system approach}
The biologist Aristid Lindenmayer \cite{ochoa1998introduction} created a simultaneous rewriting framework in 1968, which uses a set of composition laws to regulate the evolution of a character string ($\omega$). This approach is now called the Lindenmayer (or L-System) system. L-System is initially used to provide a mathematical formalism for the creation of basic complex organisms like algae and fungi, and later expanded to mimic the complex branching systems of large plants found in nature. Further applications of L-System are also found in fields such as plant computer imaging.  %While researchers worked to advance the scientific theory and to understand the general mathematics of formal languages, L-Systems sought further applications for plant computer imaging. 
We now formally define an L-system.  \\

\begin{defn}An \textit{L-system} $\textbf{G}$ is a tuple $\textbf{G} = (A, \omega, P)$ where:
\label{def:one}
\end{defn}
\begin{itemize}
\item $A$ is a finite set of symbols or variables that fall into 2 categories; (1) that can be replaced by other variables and (2) remains constant and cannot be replaced ;
\item $\omega$ is a string of symbols from $V$ that defines the initial state of the system and is called the axiom;
\item $P: A \rightarrow A^\star$ is the set of production rules that define how one variable transform into another. The predecessor is a single variable and the successor is a string from the set $A$.
\end{itemize}

%\begin{itemize}

%is a set of symbols containing both elements that can be replaced (variables) and those which cannot be replaced ("constants" or "terminals")

 %   \item an alphabet of constant and/or variable characters that is used to define the actions of the topological \cmt{(double check this)} drawing algorithm and/or assign material functionalities.
 %   \item an axiom, which is a string and provides a foundation for the full topology string of a structure. 
  %  \item a set of production rules applied first to the axiom to create a new string or strings and then recursively to any variable characters within the resulting string(s), allowing for the generation of a set of instructions corresponding to potentially complex topologies.
%\end{itemize}

The recursive nature of the L-System leads to self-similarity, therefore fractal-like and branched forms are easily generated. Plant models and other natural-looking forms are easily attained, as increasing the number of recursions causes the model to ‘‘grow” and generate a more complex self-similar structure. In our model, we start with an axiom and define a set of rules to describe branch structures as seen below: \\
\\
$Variables: \  A = \{g, d\}$

$\{Axiom: \ \omega = \{g\}$

$Rule 1: \ P(g) \rightarrow d(d)+d)[d(d)+d)$  % 4 branches

$Rule 2: \ P(g) \rightarrow d(d)+d)[d(d)+d)[d(d)+d)$  % 6 branches

$Rule 3:  \ P(g) \rightarrow d(d)+d)[d(d)+d)[d(d)+d)[d(d)+d)$\\  % 8 branches

As seen above, we have 2 variables, where the variable $d$ is a constant and our axiom is the single character $g$. It is to be noted that different rules give rise to different branching patterns. For example, Rule 1 give rise to a tree with 6 branches and Rule 2 corresponds to a tree with 12 branches. This L-system can easily be extended and generally applied to create branch patterns. In Figure \ref{fig3} (a), we demonstrate the plot of a simple tree with one level branching structure. As shown in Figure \ref{fig3} (a), L-system only provides simplified branching structure. In order to simulate natural looking trees, more characteristics such as curvature of the branches need to be added. We achieve this by modify geometric information contained in CAD developed files.

\subsection{The CAD developed approach}
Computer-aided design or CAD is a way of creating models of objects using computers. CAD software can be used to design both 2D and 3D object models, producing digital files that contain different parameters of the models. For example, a 3D model designed using CAD contains all details required to make a 3D print of the model. The 3D model is generally shown as a 2D image with 3D rendering or visualization. It is worth to note that, although the term 3D model refers to the object,
the 3D file refers to both the object and the corresponding file type. A SolidWorks file and an STL file, for example, are both 3D files and can represent the same 3D object/model, but they do so differently and are thus different 3D files. 

Regarding the representation, 3D models are often represented using the following formats: (1) a solid model that defines the volume occupied by the object, or (2) a surface representation model that models the surface of an object.  Solid models are generally used for engineering analysis, for example, in limit load and failure analysis. Surface models, on the other hand, are commonly used in computer games and movies. An example of a surface model is an STL (STereoLithography) file. An STL file describes mesh surfaces as lists of geometric features. STL files contain polygonal meshes defined by vertices and normal vectors that represent the surface features of the 3D model. A solid surface is modeled as a composition of triangular faces, and each triangular face comprises a normal and the coordinates of the 3 vertices that form the triangle. The STL file stores information mostly in the binary form although a text representation can also be used.  

%\cmt{In Section II A and B, you only reviewed L-system and CAD files. You did not mention your approach, i.e., how you combined these to simulate random trees. I suggest you add one paragraph in each section to describe how you used these to simulate a random tree.}

In the first step, the tree with adequate characteristic was chosen. The information obtained from the measurement of 3D CAD model has been used to visualize the object files of the trees in freeCAD software. An example is shown in Figure \ref{figcad1}. The information about branches, sub-branches and leaves can be first extracted based on the coordinates and the meshes information from the object files. In second step, this information enables the placement of branches in accordance with the rules of L-system branching pattern. Since the coordinates of branches are available, we can randomize the angles and these geometry of these branches. The sub-branches can be added and modified by following the same steps. The leaves are then added by including the triangular meshes of leaves from object files, as shown in \ref{figcad2}. The center point of each triangle is considered and randomized as per the change in angle of sub-branches.

%The random placement of the leaves along the sub-branches and branches has been done.

%Further, in the second step, the sub branches has been developed by using the same meshes and faces information. Figure \ref{figcad2} shows the constructed model compared to the real tree. The model different from the real plant in the little amount of angle and density between leaves. 
 %This enables random placement of these coordinates along the tree trunk. The randomization can be uniform, gaussian etc.

% The answer is the paragraph by the end of IPP, will arrange it properly

\subsection{Inhomogenous Poisson process}
%Non-homogenous Poisson processes represent a group of non-stationary point patterns. The only difference between this method and the homogenous Poisson process is that the density $\lambda$ is no longer constant within the defined region $\Re$, it becomes substituted by a location dependent density function $\lambda(X)$.

%Inhomogeneous Poisson point processes (IPPPs) are versatile stochastic pro-cesses that describe points or ”events” that occur with a changing rate and that are stochastically independent from each other. IPPPs can be used to modela variety of spatial and temporal processes, such as the numberof customersarriving at a supermarket as well as the time or the location of animalsightings.The aim of this paper is to present a number of numerical approaches to gener-ate random numbers corresponding to the location of points or thenumber ofpoints in a set generated by an IPPPs. For the cases where additional informa-tion such as the locations of points or the number of points occurring is availableupfront, procedures to derive the conditional probability densities are given inaddition to the procedures to generate the corresponding random numbers.

Inhomogeneous Poisson Processes (IPPs) are flexible stochastic processes that model random points in space or random "events" in a time interval. % It is characterized by a Poisson process, which is a stochastic process that is simple and commonly used to model the times when arrivals enter a system. 
IPPs can be used to model a multitude of spatial and temporal phenomena, such as the number of cars passing through a junction or the timing/place of animal sightings. Based on a random tree simulator, we are able to generate a forest consisting of a random number of trees. By sampling from the IPP, we can determine the number of trees as well as the positions of these trees in a 2-D region. 

Specifically, let $D \in \mathbb{R}^2$ denote the region in which the group of trees will be built. The random locations (i.e., $(x, y)$ coordinates) of the trees will be denoted by $S = \{s_i, i=1, \dots, n\}$. We assume that S follows an IPP with intensity function $\lambda(s) : D \rightarrow R^{+}$, where $\lambda(s)$ is a parameter that control the tree density on $D$. Small values of $\lambda(s)$ indicate sparse regions whereas high values indicate dense regions. Given the region $D$ and the intensity $\lambda(s)$, the number of trees, $n$, follows a Poisson distribution with mean $\int_D  \lambda(s) ds$. To simulate $S$ given $n$, we adopt a thinning approach, and details can be found in \cite{lewis1979OR}.

%\cmt{Somewhere in Section II you should add a description about how to add randomization to CAD files.}

% The randomization in CAD files and also in Lsystem has been added in simulation enviroment by using the rand funtion of MATLAB.

\section{Simulation Results}
\begin{figure*}[h]
    \centering
        \subfloat[]{\includegraphics[scale=0.29]{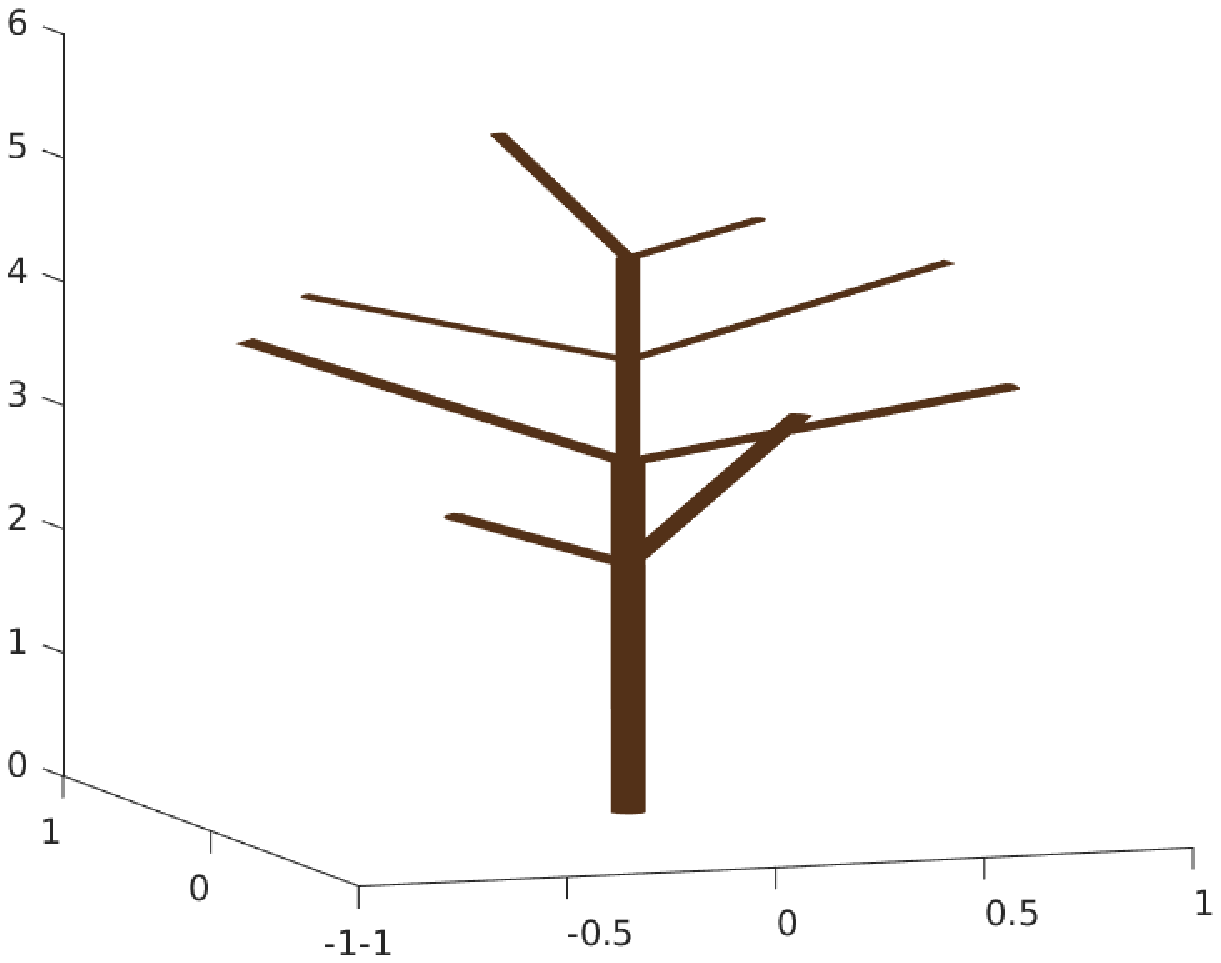}}
\subfloat[]{\includegraphics[scale=0.29]{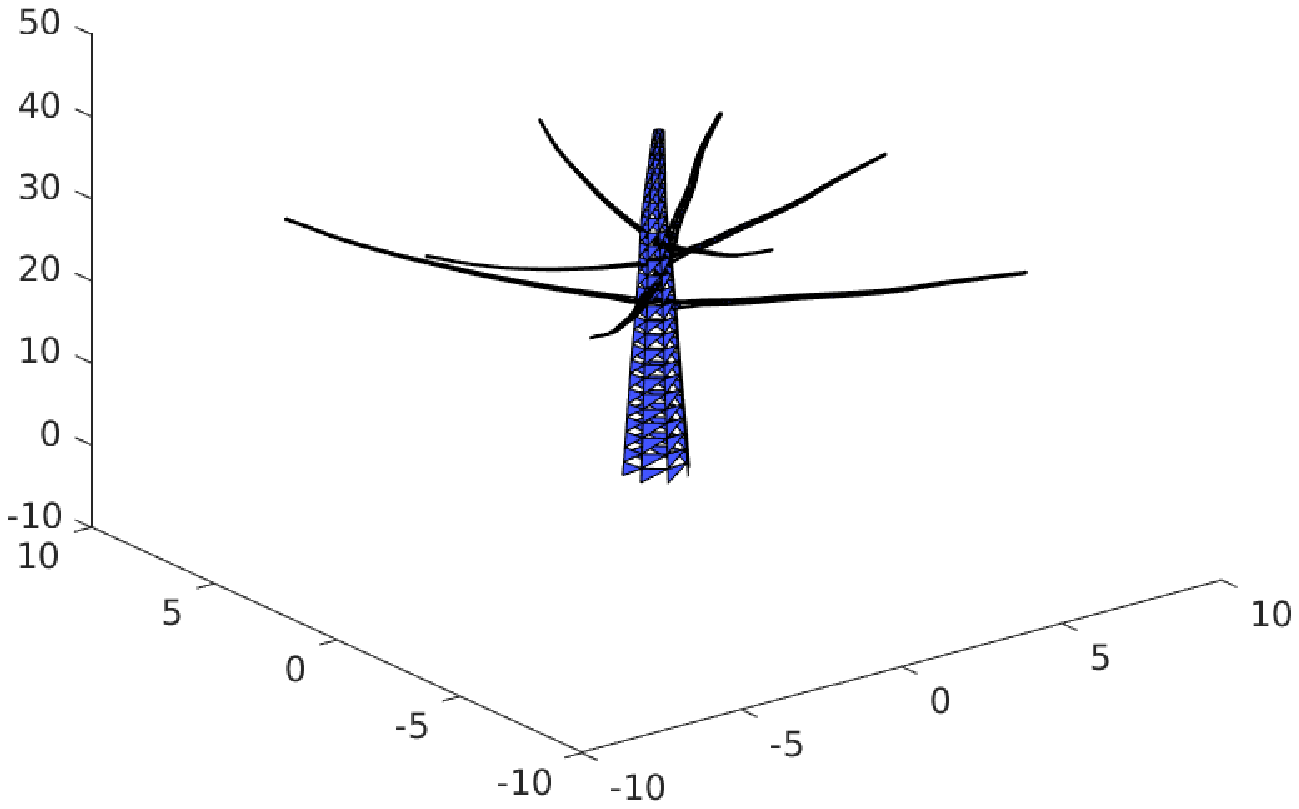}}
\subfloat[]{\includegraphics[scale=0.29]{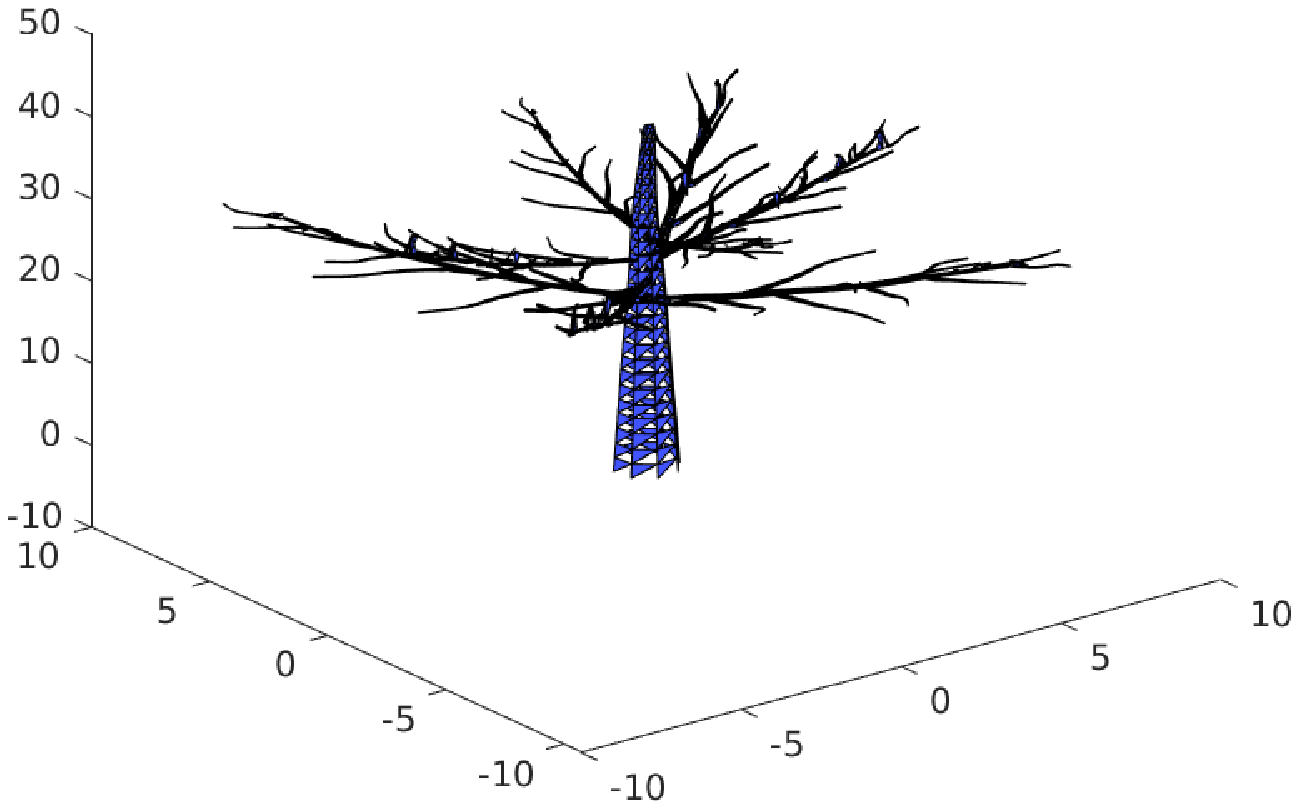}}
\subfloat[]{\includegraphics[scale=0.29]{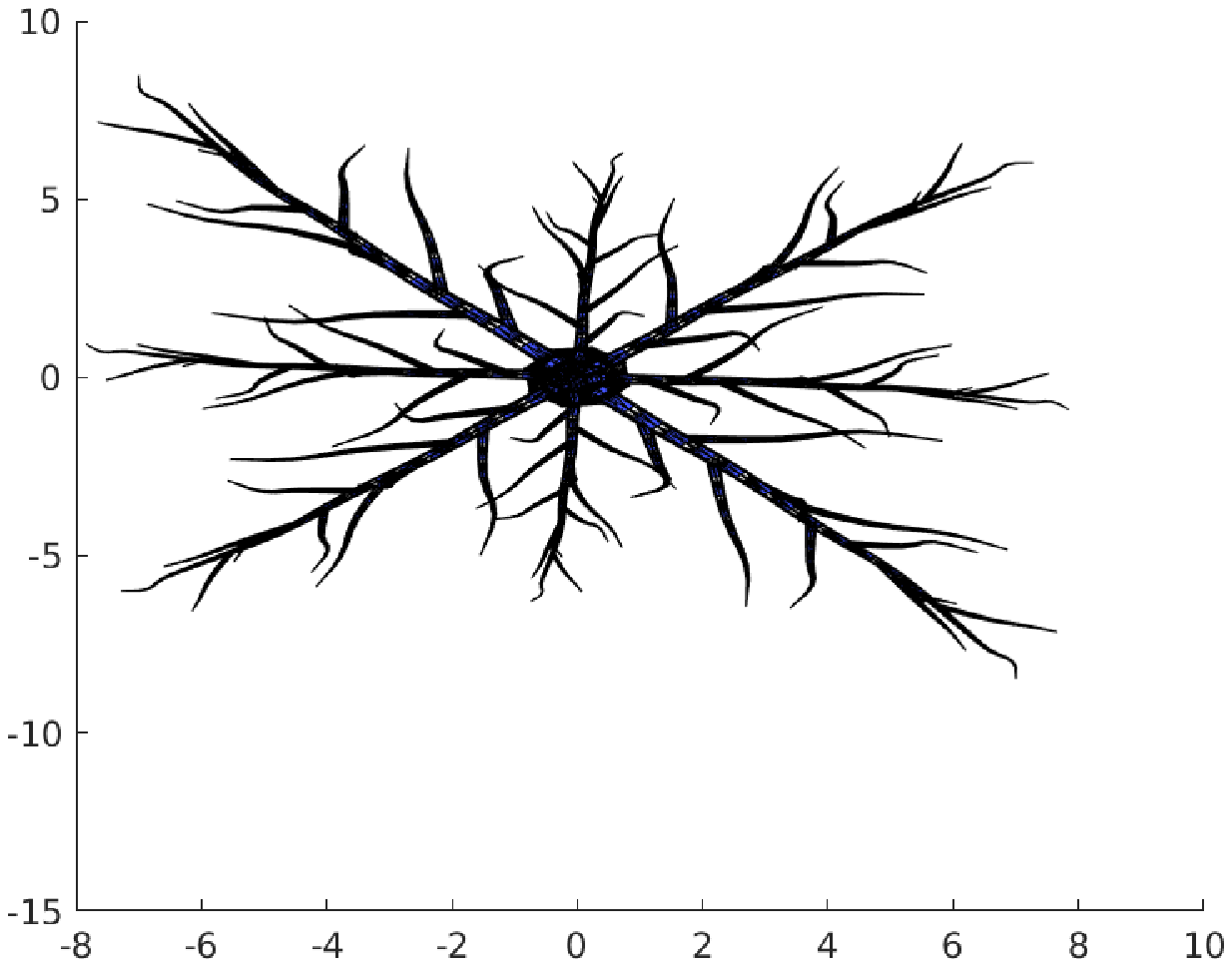}}
\caption{Different stages of simulating a tree. (a) 8 different branches generated according to L-system rule. (b) The L-system model is used as a base to create a CAD, to mimic naturally occurring branches. (c) Sub-branches are added. (d) A top view of the final tree without leaves. The trunk, branches and sub-branches are modelled using triangular meshes.}
        \label{fig3}
\end{figure*}

\begin{figure*}[h]
    \centering
        \subfloat[]{\includegraphics[scale=0.29]{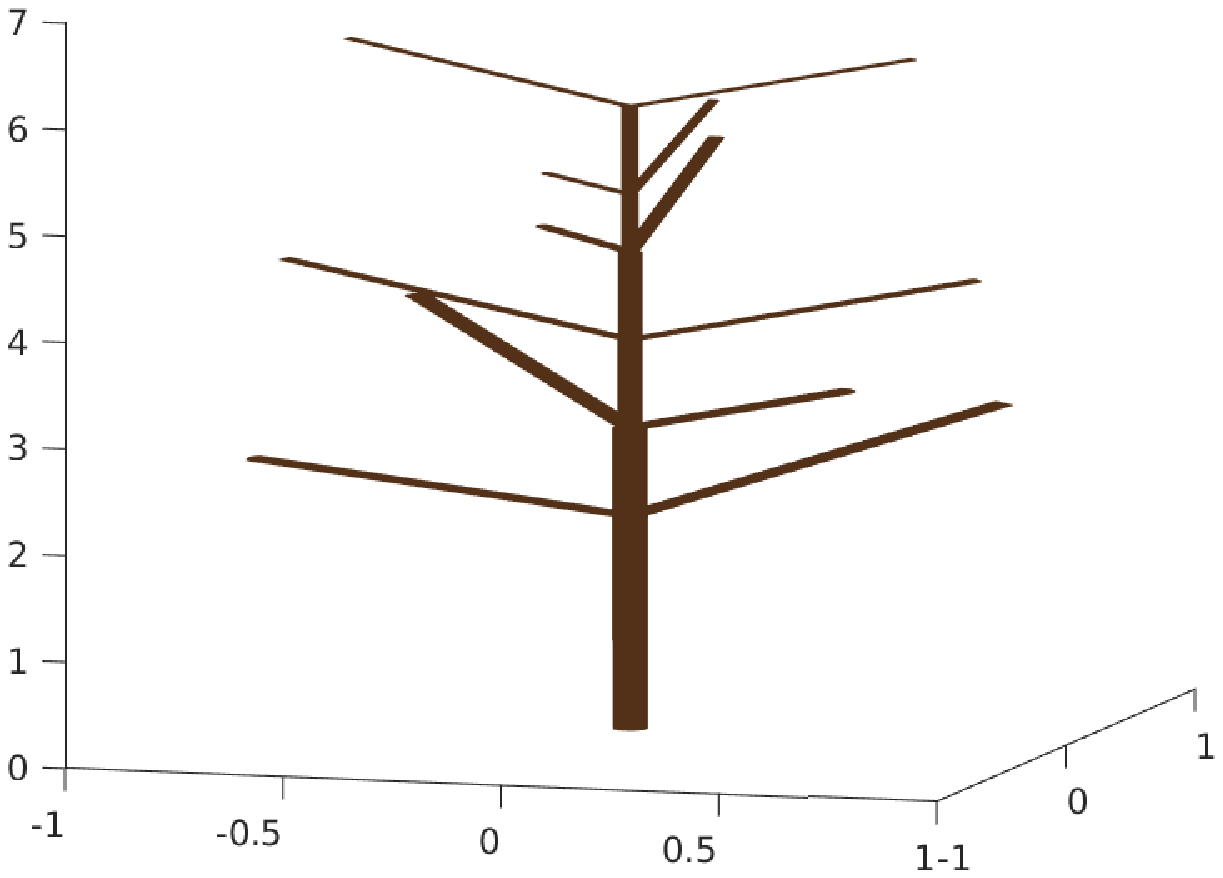}}
      % \subfloat[]{\includegraphics[scale=0.29]{lsys1_top.eps}}
\subfloat[]{\includegraphics[scale=0.29]{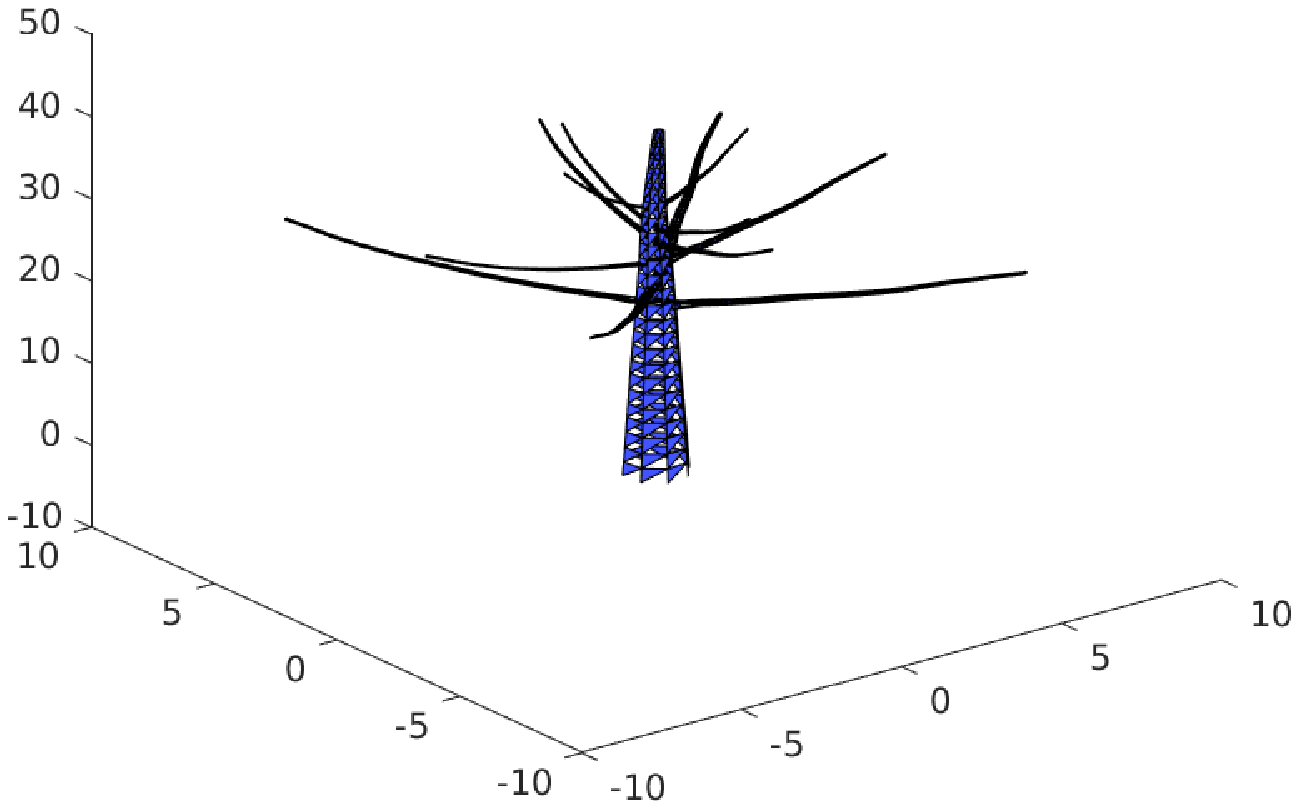}}
%\subfloat[]{\includegraphics[scale=0.29]{lsys1_3D_branch_top.eps}}\\
%\subfloat{\includegraphics[scale=0.29]{lsys1_xy.eps}}
%\subfloat{\includegraphics[scale=0.29]{lsys1_3D_branch_side1.eps}}
\subfloat[]{\includegraphics[scale=0.29]{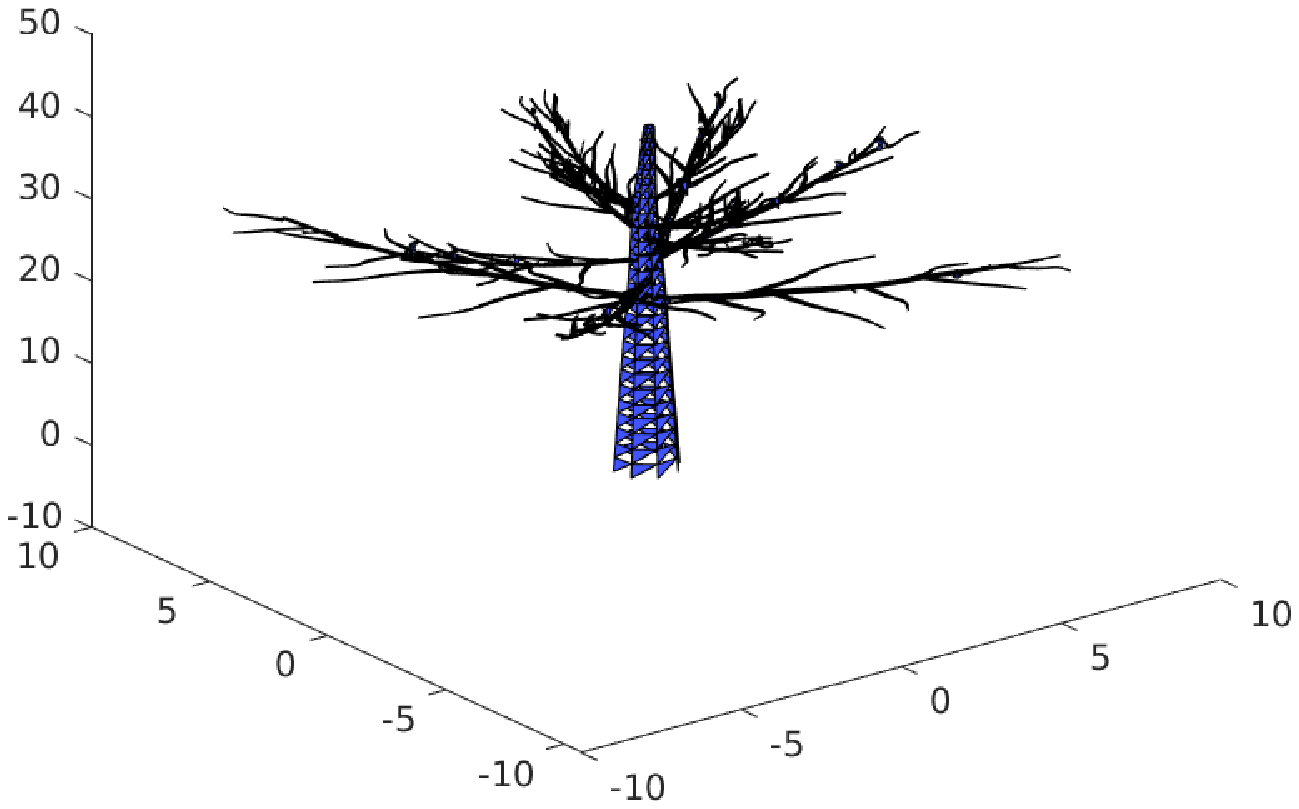}}
\subfloat[]{\includegraphics[scale=0.29]{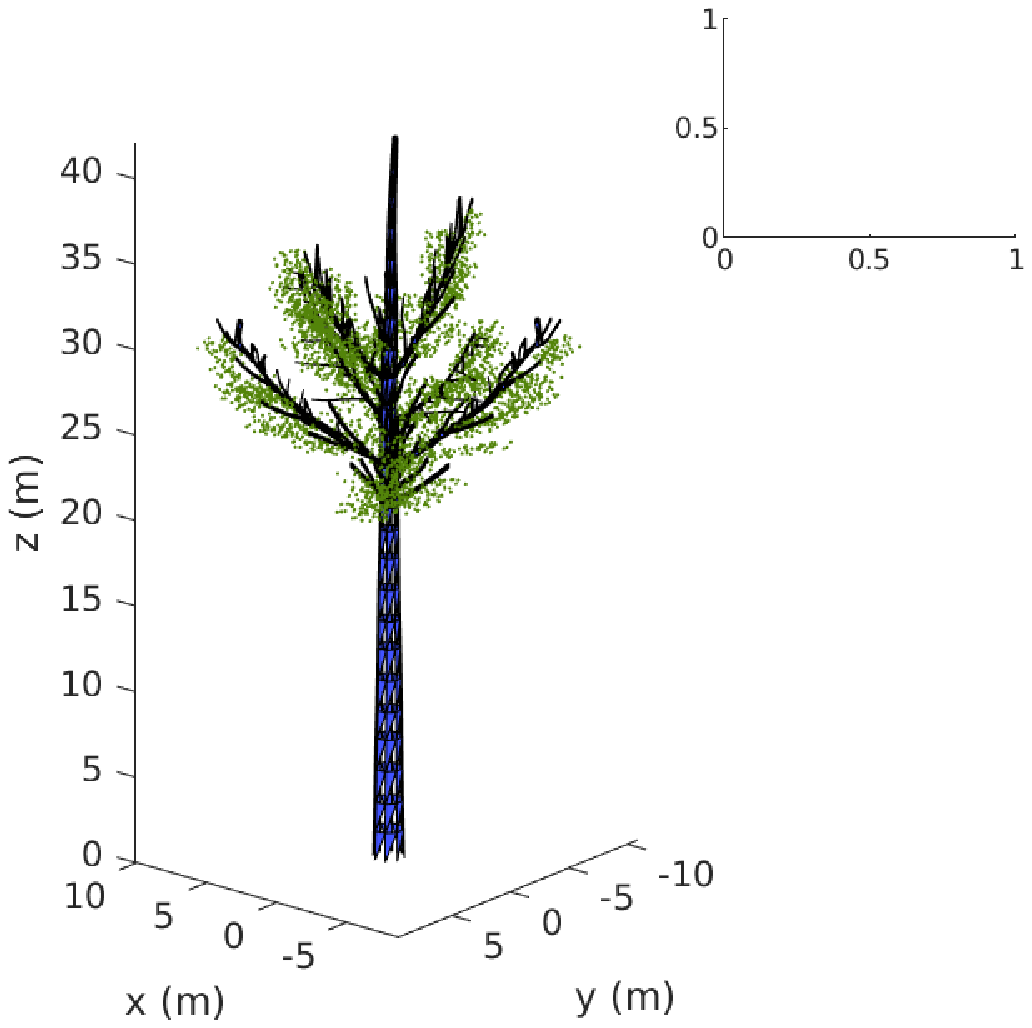}}
\caption{Simulating a tree with 12 branches. (a) The branch locations and their orientation with respect to the trunk is seen. The branching pattern is in accordance with L-system rule 2. (b) 3D CAD model that mimics the L-system branching pattern. (c) Sub-branches added to the model. (d) Shows the complete 3D model with the leaves. Leaves are modelled as triangular meshes, but only the midpoint of the meshes are shown (green) for better visualization.}
        \label{fig1}
\end{figure*}

The primary intent of  this  study is develop a stochastic model to simulate random trees. We achieve this by integrating L-systems with CAD developed  analytical tree models. In this section, we demonstrate results from several scenarios.  %In general, the results  of these models have been found to hold  under stochastic conditions. In particular, it has been shown that the performance of a our method is  equivalent to that of the closest natural environment.

The first scenario is demonstrated in Figure \ref{fig3}. In this scenario, we simulate a tree with only eight branches. The starting points of the first level branches have been simulated using the L-system. The geometry about the branch lengths and sub-branches is obtained from the 3D developed CAD models.

The second scenario is demonstrated in Figure \ref{fig1}. In this scenario, we simulate a random tree with 12 branches. The starting points of were simulated by using L-system. The branch lengths and sub-branches were then simulated by adopting the geometry of 3D developed CAD models. In this case, we added information about leaves, which is obtained by using the same geometry information of CAD model as shown in Figure \ref{figcad1}. Each leaf is a triangular mesh, but only the center points of the leaves have been plotted.

\begin{figure*}[h!]
    \centering
        \subfloat{\includegraphics[scale=0.29]{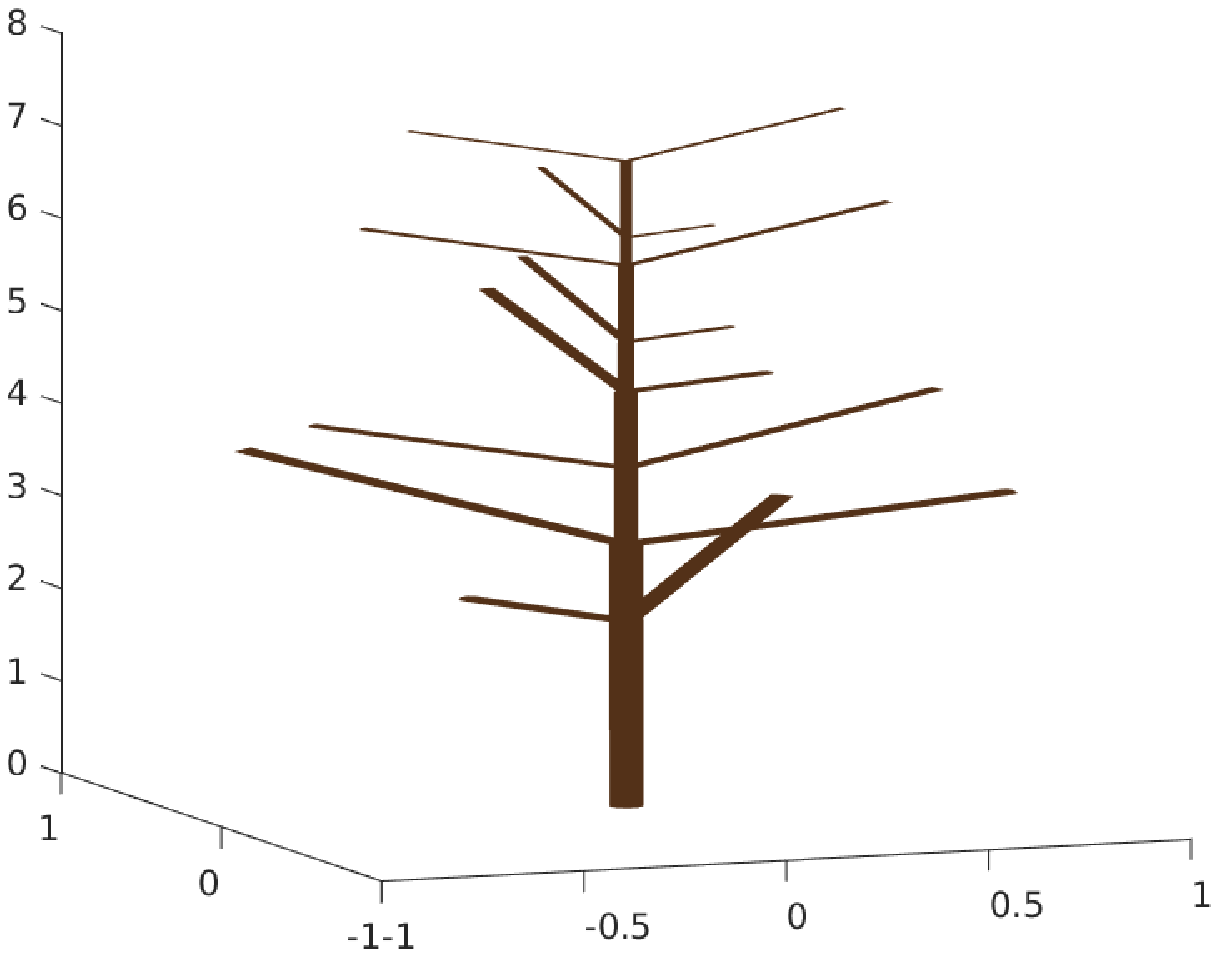}}
\subfloat{\includegraphics[scale=0.29]{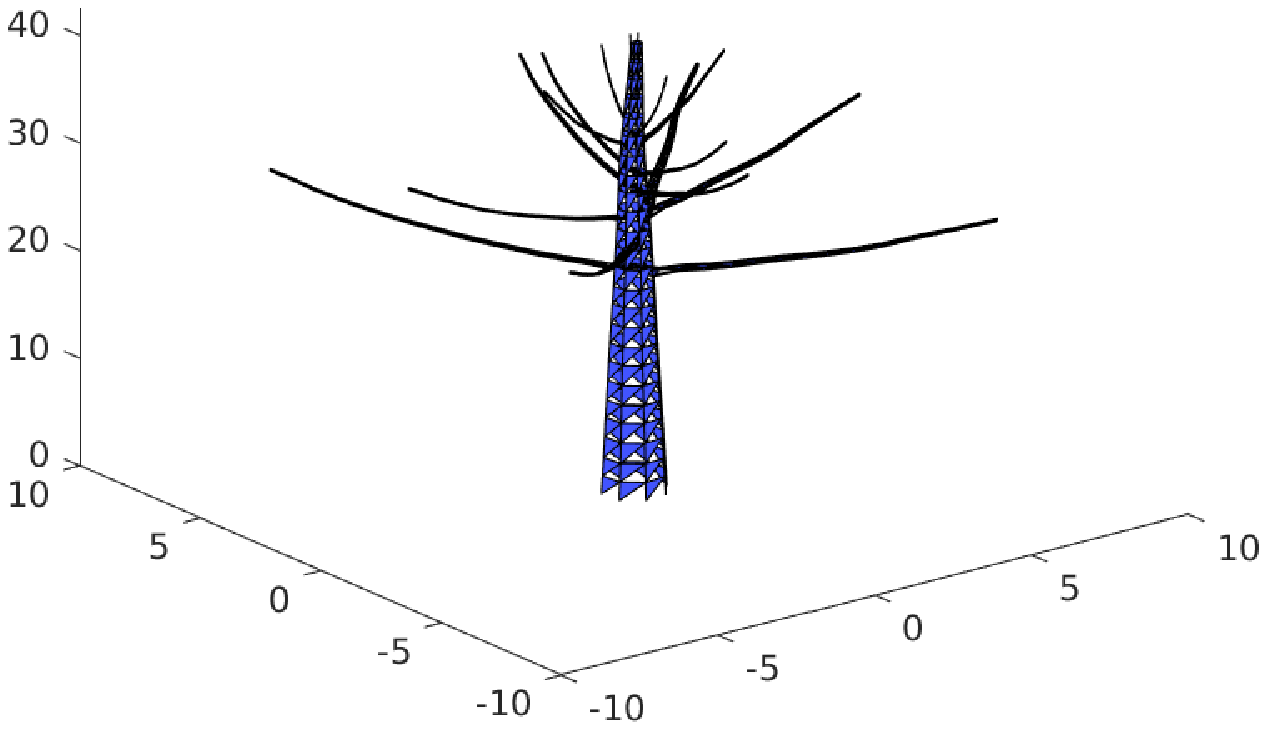}}
\subfloat{\includegraphics[scale=0.29]{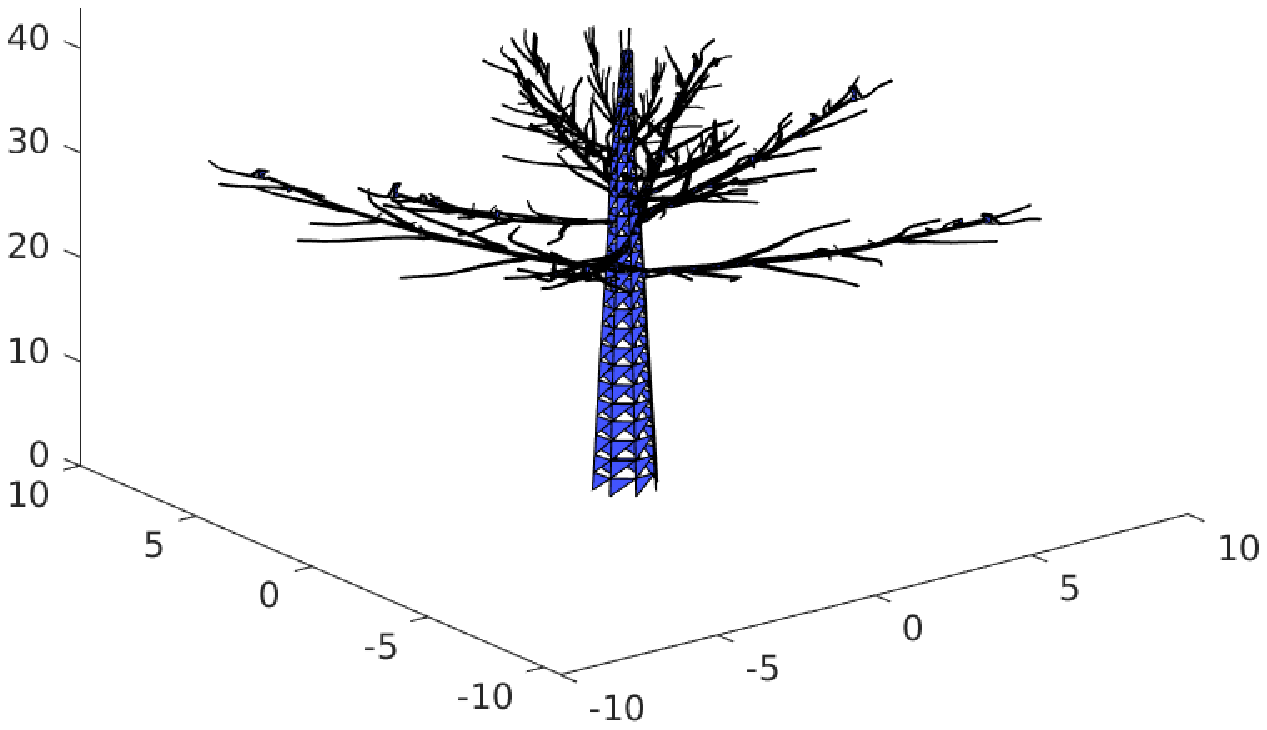}}
%\subfloat{\includegraphics[scale=0.29]{lsys2_3D_subbrach_top.eps}}
\subfloat{\includegraphics[scale=0.29]{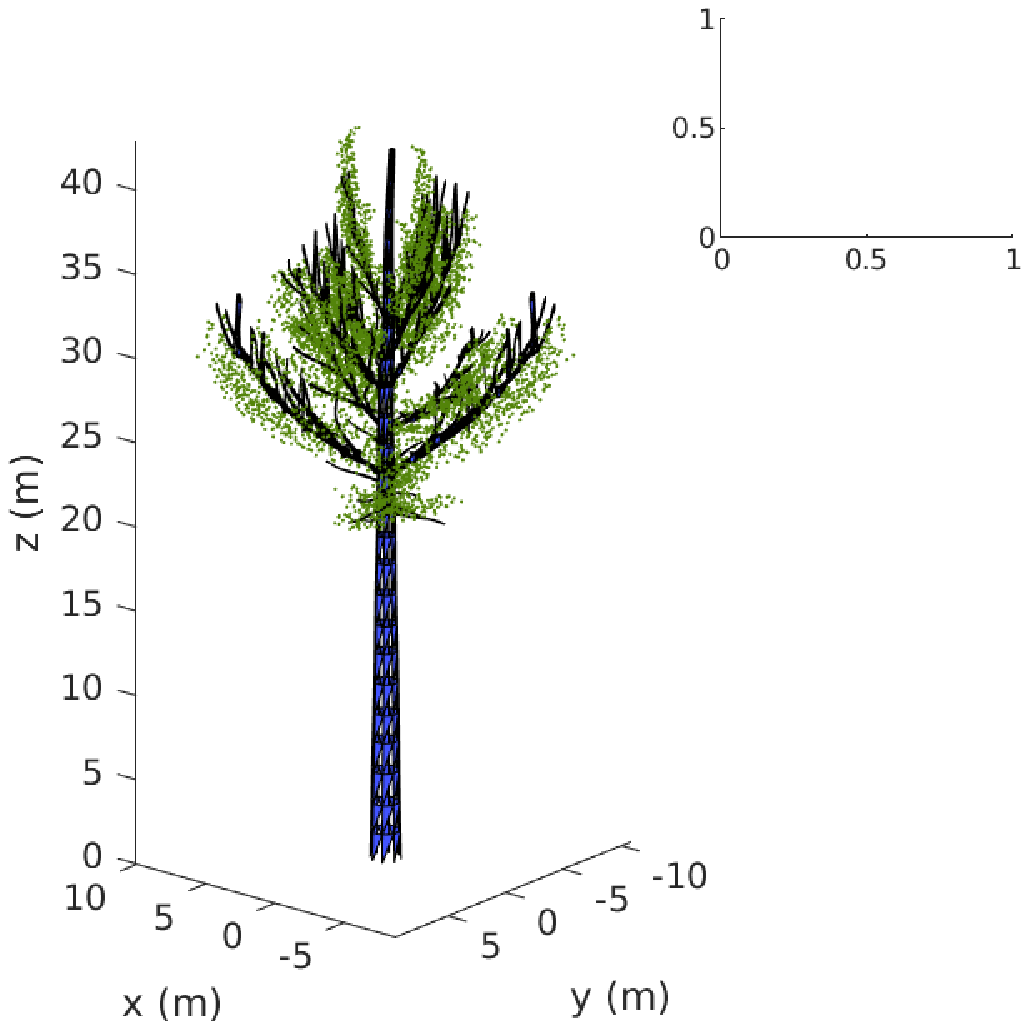}}

        \caption{Different stages of simulating a tree with 16 branches.}
        \label{fig2}
\end{figure*}

We show the third scenario in Figure \ref{fig2}, which is similar to scenario 2, but with 16 branches simulated.  

\begin{figure*}[h]
    \centering
      %  \subfloat{\includegraphics[scale=0.31]{IPP.eps}}
%\subfloat{\includegraphics[scale=0.31]{IPP1.eps}}
\subfloat{\includegraphics[scale=0.27]{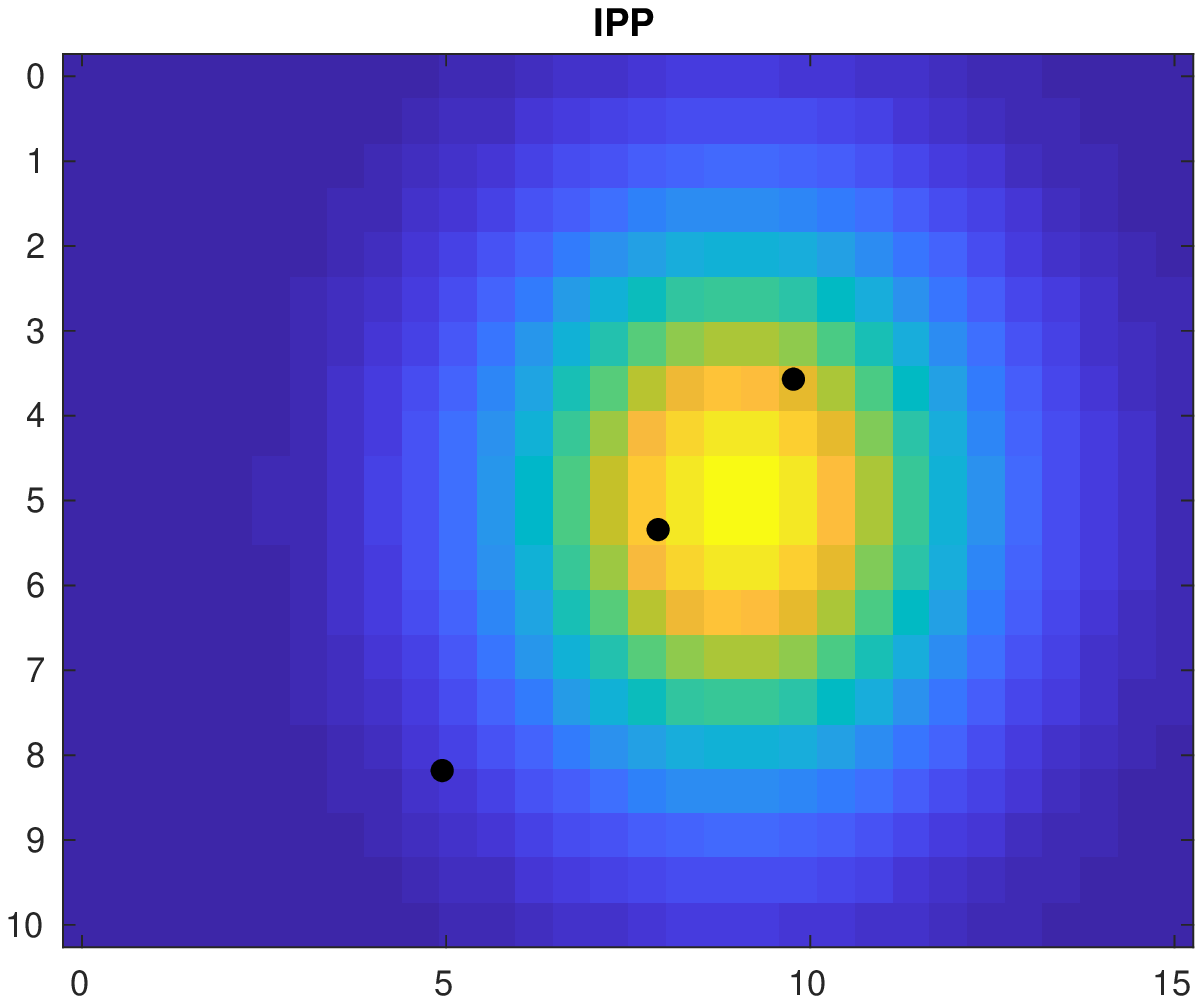}}
\subfloat{\includegraphics[scale=0.27]{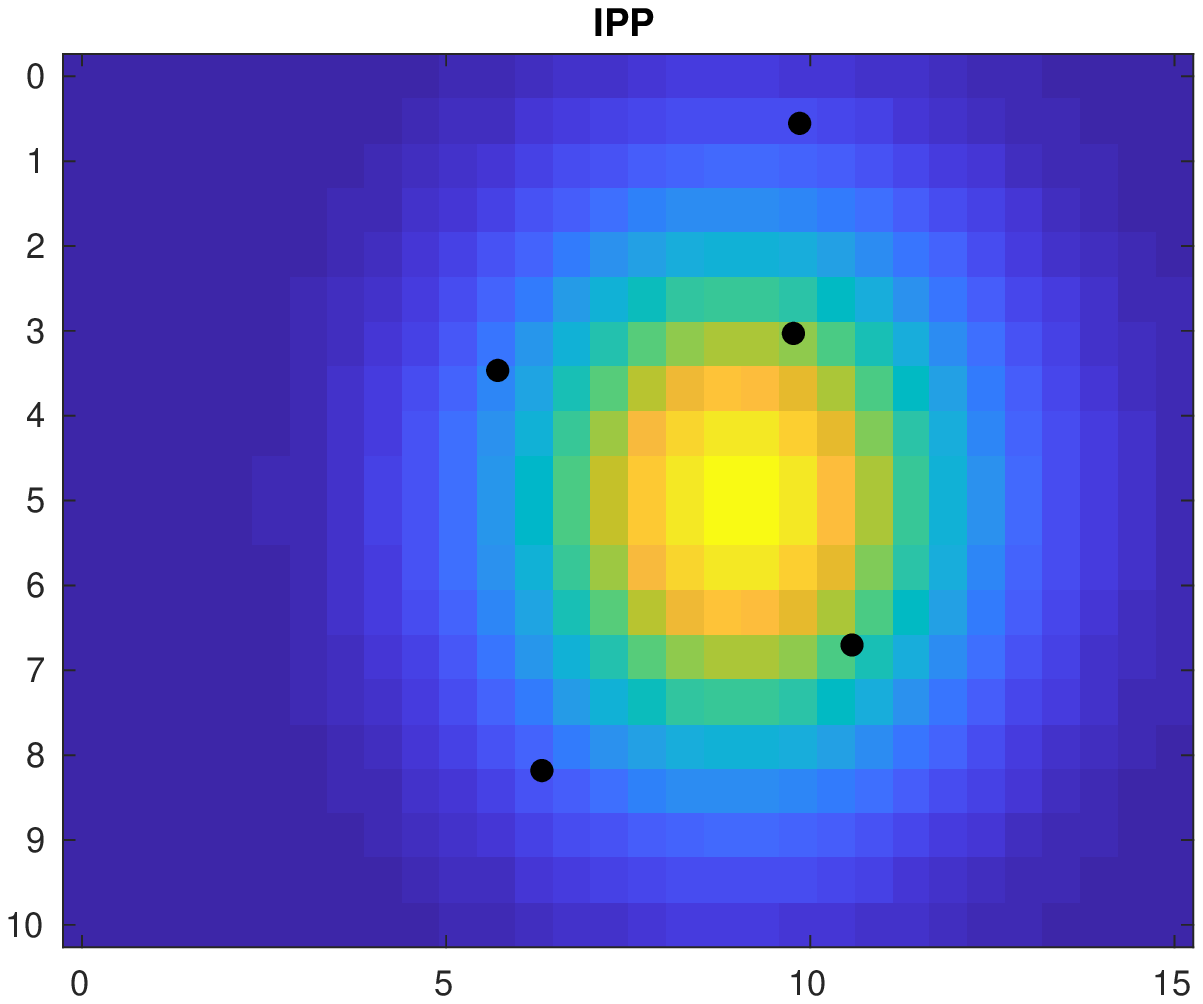}}
\subfloat{\includegraphics[scale=0.27]{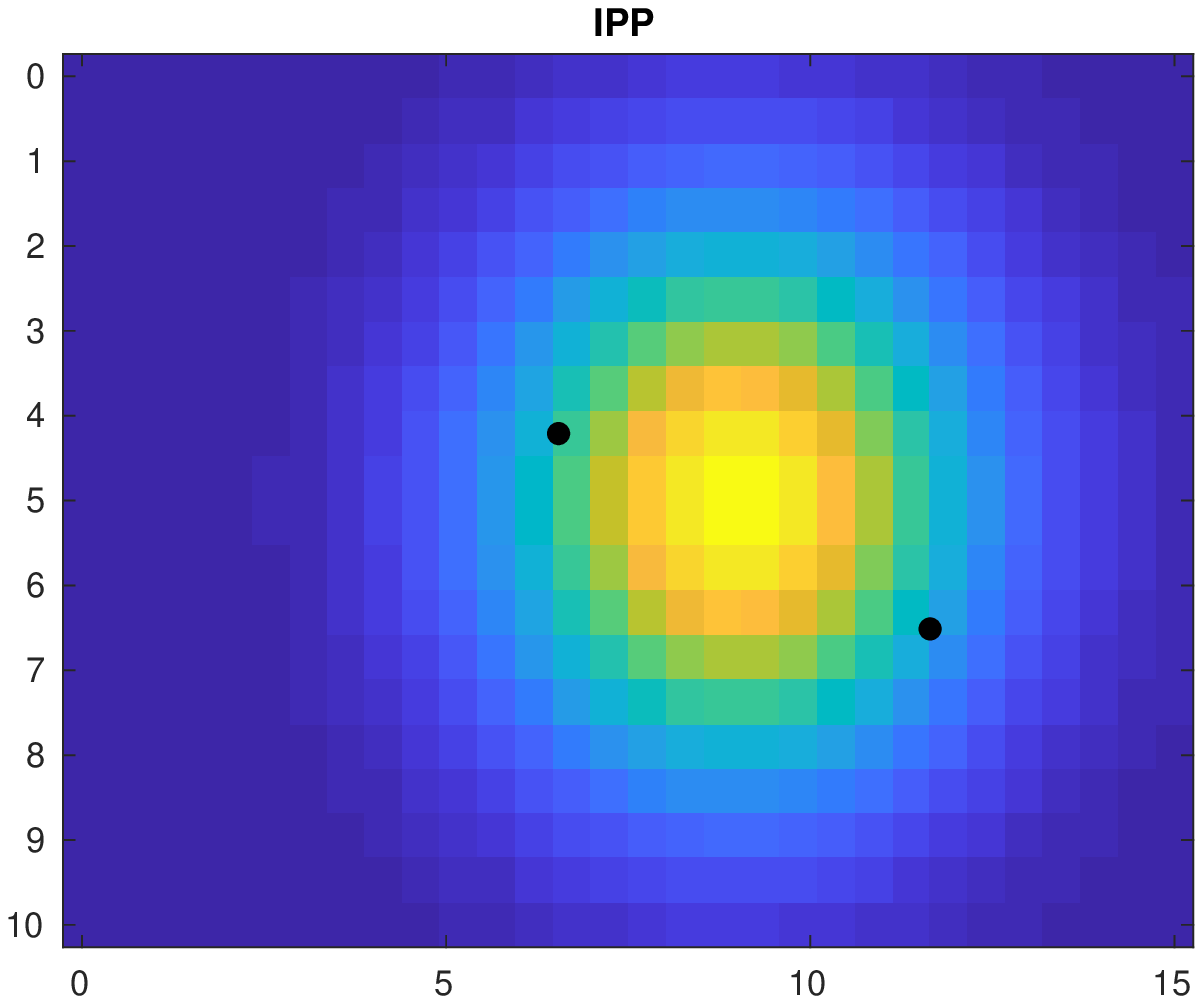}}
%subfloat{\includegraphics[scale=0.27]{IPP5.eps}}
\subfloat{\includegraphics[scale=0.27]{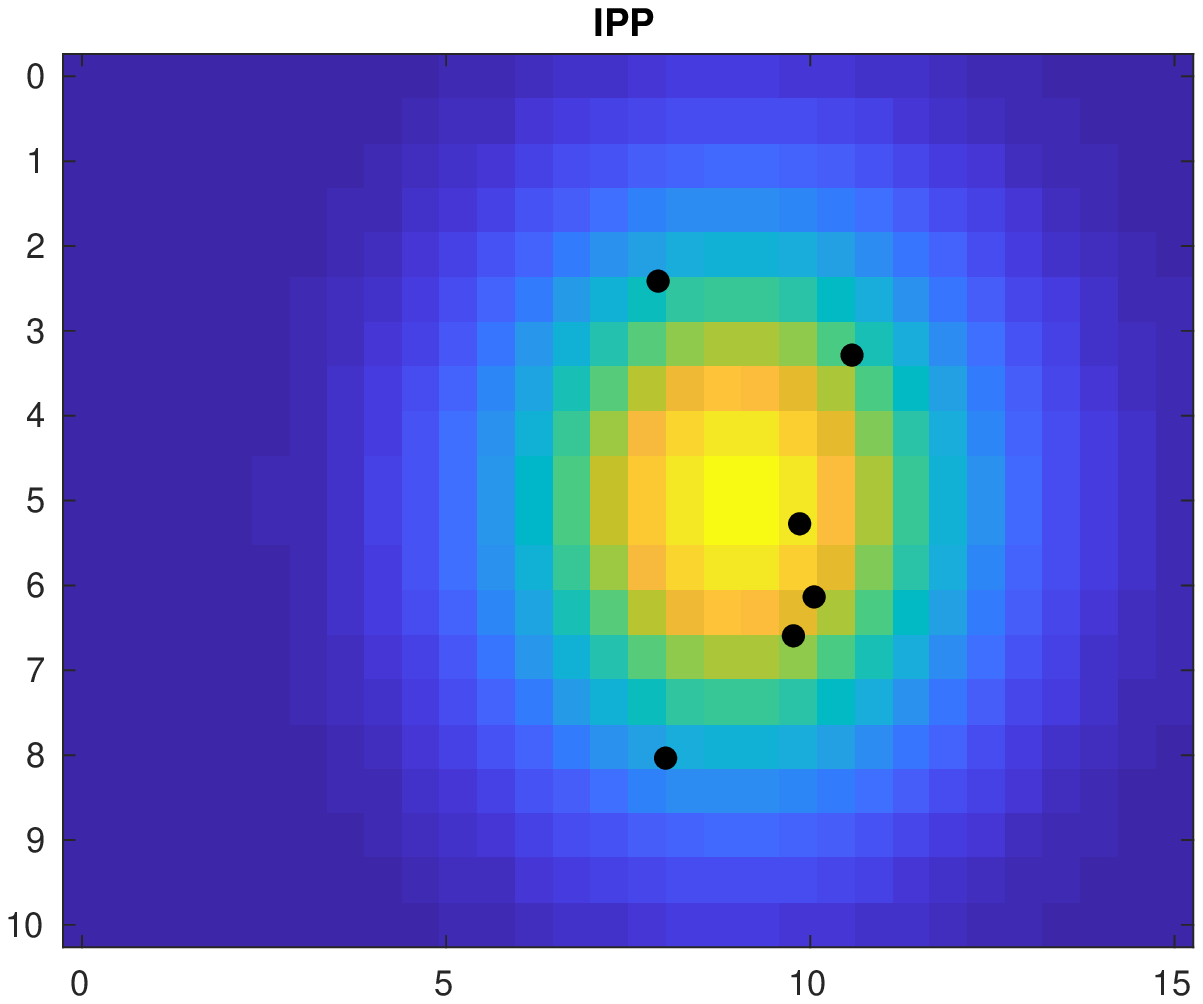}}
        \caption{Different number of trees sampled in a region to create a forest. The tree locations are sampled from an IPP. The number of trees in a region depend on the IPP parameter.}
        \label{fig4}
\end{figure*}

The above results demonstrate that it is feasible to simulate random trees with different number of branches, sub-branches and leaves. To simulate a forest with multiple trees, an IPP approach has been considered. Figure \ref{fig4} shows the simulated locations of trees sampled from an IPP. Based on this results, we can generate multiple number of trees and plot them in a 3D environment.

\begin{figure*}[h!]
    \centering
        \subfloat[]{\includegraphics[scale=0.29]{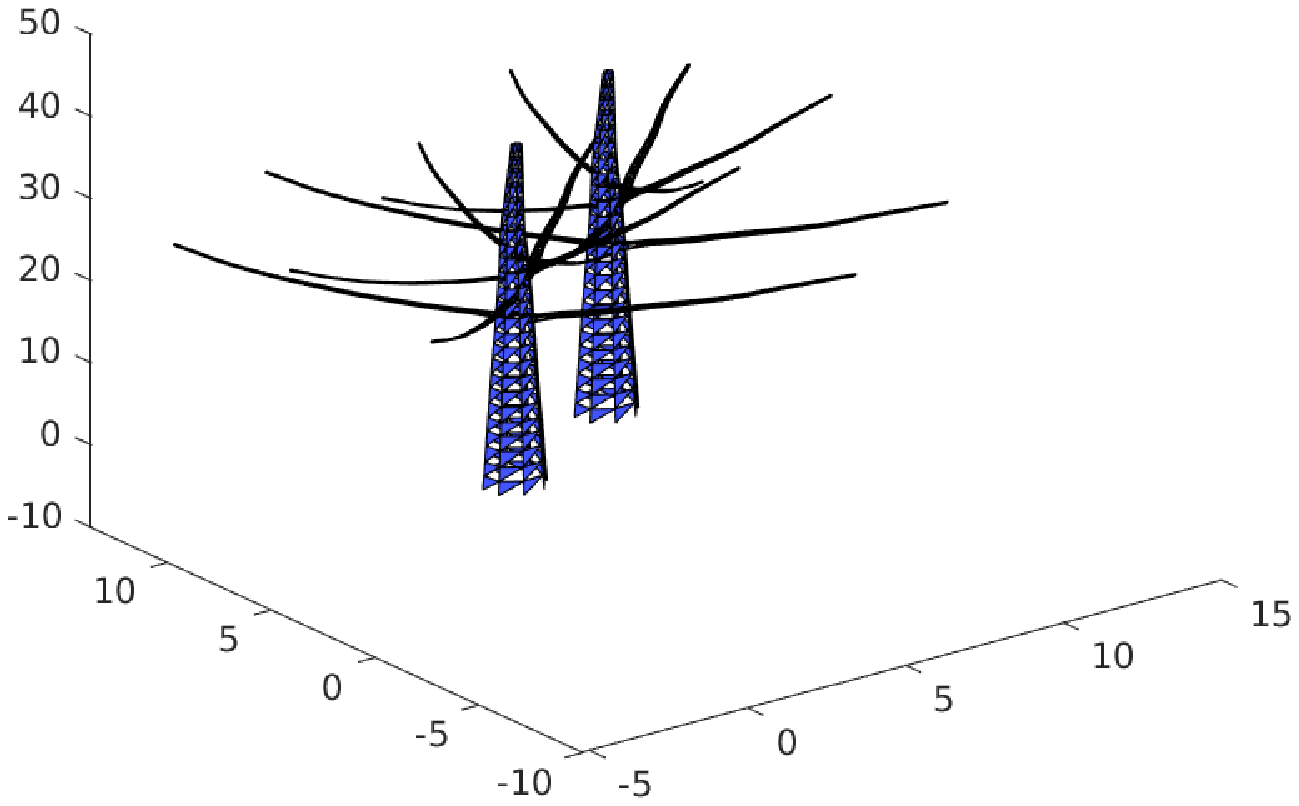}}
%\subfloat[]{\includegraphics[scale=0.35]{law3with3D_top.eps}}
\subfloat[]{\includegraphics[scale=0.29]{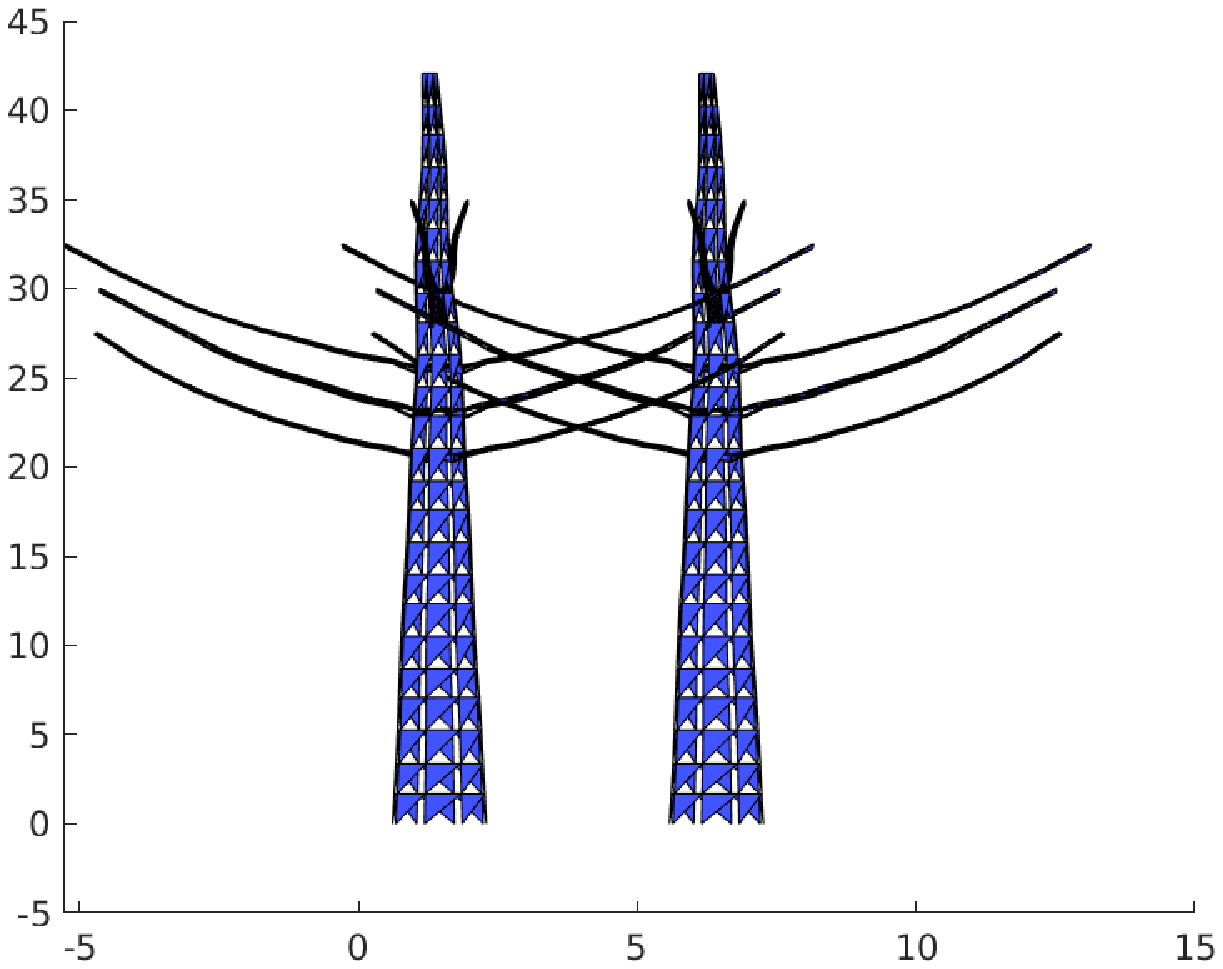}}
\subfloat[]{\includegraphics[scale=0.29]{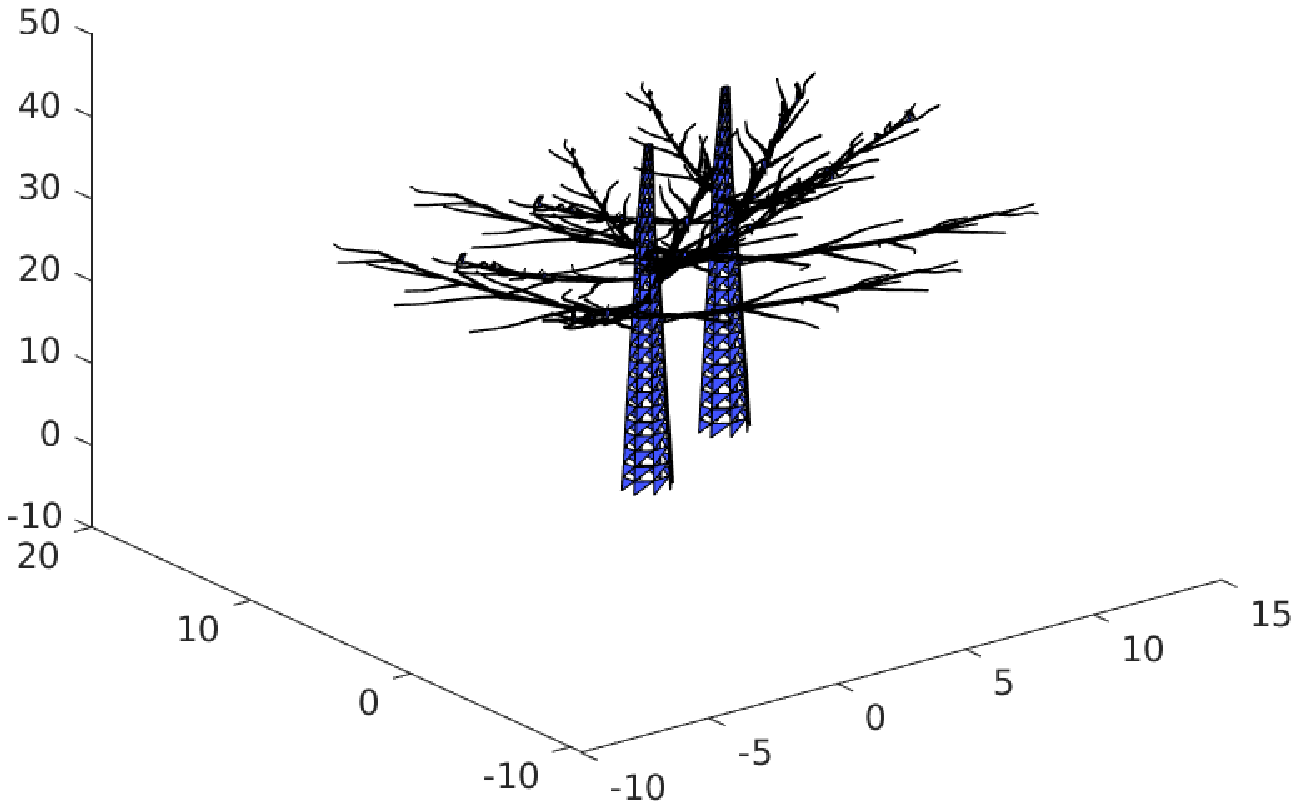}}
%\subfloat[]{\includegraphics[scale=0.35]{law3with3D_subbrancch_top.eps}}
\subfloat[]{\includegraphics[scale=0.29]{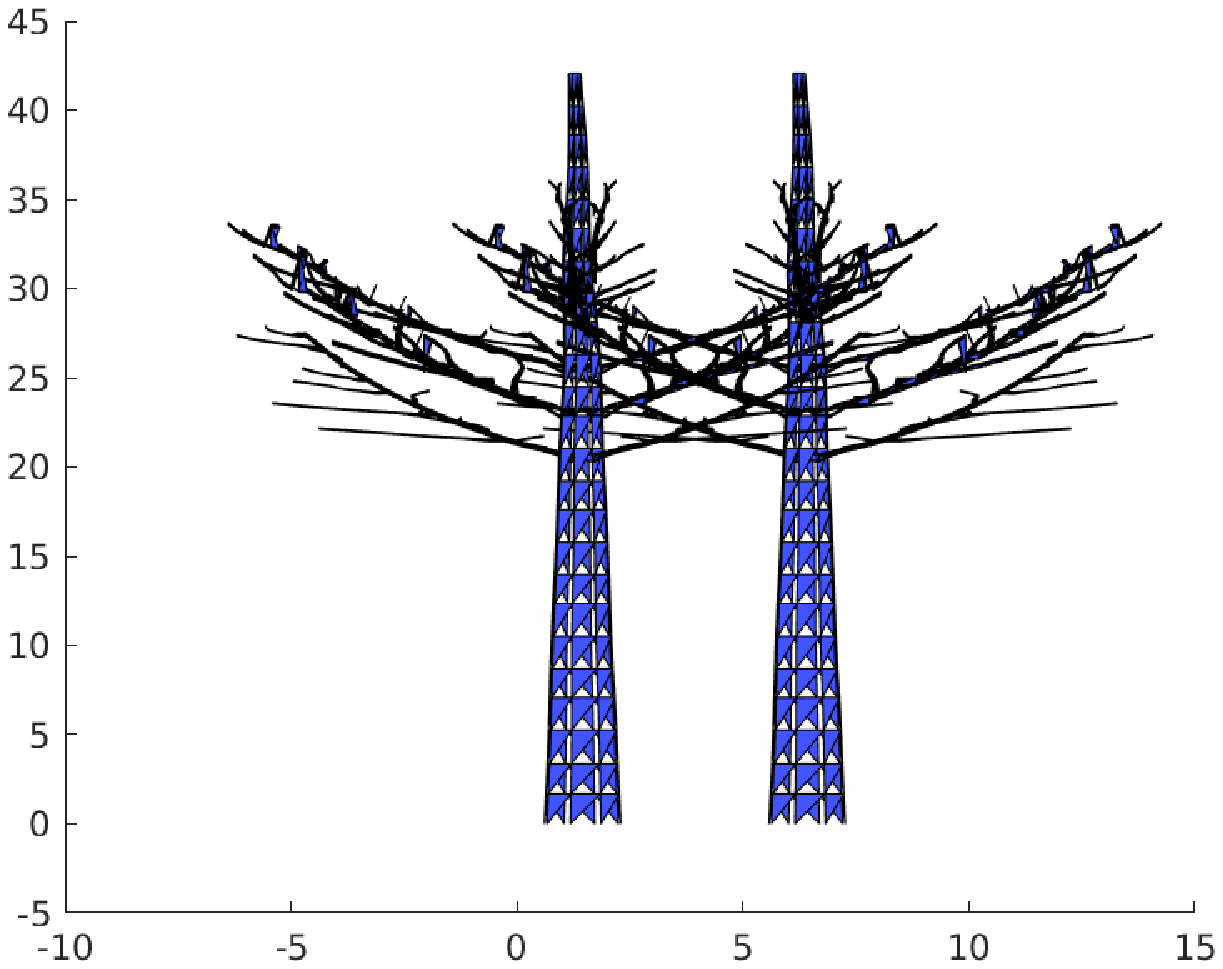}}
        \caption{Simulating 2 trees in a region. (a)-(b) 3D model of the 2 trees with trunk and branches and their corresponding side view. (c)-(d) Sub-branches are now added and the side view are also visualized.}
        \label{fig5}
\end{figure*}

We further simulated forests with multiple trees. Results are plotted in Figure \ref{fig5}.

%begin{figure*}[h]
%\centering
  %      \subfloat{\includegraphics[scale=0.40]{5tree3D.eps}}
%\subfloat{\includegraphics[scale=0.4]{5tree3D_side.eps}}
%subfloat{\includegraphics[scale=0.4]{5tree3D_top.eps}}\\
%subfloat{\includegraphics[scale=0.4]{5tree3D_subbranch_3D.eps}}
%\subfloat{\includegraphics[scale=0.4]{5tree3D_subbranch_side.eps}}
%subfloat{\includegraphics[scale=0.4]{5tree3D_subbranch_top.eps}}
     %   \caption{CAD generated model for 5 different trees in an environment. Notice that all the trees have variations in the lengths of their branches and sub-branches. Their angles with respect to the tree trunk also vary from tree to tree. This variation in pattern is incorporated by randomly sampling the quantities from a uniform distribution.}
    %    \label{fig6}
%end{figure*}

These plots demonstrate the effectiveness of our approach in generating natural looking trees in simulation environment. As many other 3D developed tree models are available online, our approach can be used to simulate a large community with different  tree species. This work can be used in any application when a simulation model is needed for the testing of algorithms. For instance, one can simulate
the motion of a robot in various scenarios such as obstacle avoidance and path following.

\section{Conclusion}
%In this paper, we have developed a new way of creating and analysing 3D trees in simulated environment. The fusing of 3D develop object files and L-system makes it possible to build and manipulate geometric models of trees and their parts. Our method can easily be extended and applied to other problems in 3D modeling that require more faithful performance. 

We have developed a new way of simulating and analysing 3D trees in simulated environment. By integrating the branching patterns generated by L-systems and the geometries of 3D CAD developed object files, our approach results in natural looking trees with full geometry about branches, sub-branches and leaves. Our approach also provides a way to simulate a forest by sampling the random locations of trees from an IPP model.  

Our approach can be used to create complex structured 3D virtual environment for the purpose of testing new sensors and training robotic algorithms. One of the most prominent application of this research is the study of smart sonar sensors that mimic the biosonar system of bats. More than 800 species of echolocating bats use ultrasonic waves to detect their surroundings in forest. The highly efficient sonar systems of bats allow them to detect objects as thin as human hairs in complete darkness. They also enable them to efficiently navigate through dense vegetation by using echoes. To learn these impressive abilities from bats, it is imperative to simulate foliage echos data in a dense natural environment. Simulating real world trees is the first step towards achieving efficient navigation in forest. Future work, can be an extension of this work in robotic applications where a drone is flying with on board monocular camera and knowing the position of foliage using Inverse perspective mapping \cite{10.1007/978-3-030-31993-9_21} and hence generating echos.

\FloatBarrier

\section*{Acknowledgment}
This research is supported in part by National Science Foundation (NSF) grant \#1762577.
%\section*{References}

\bibliography{references}
\bibliographystyle{unsrt}

\end{document}